\documentclass[sigplan,10pt]{acmart}
\AtBeginDocument{%
  \providecommand\BibTeX{{%
    \normalfont B\kern-0.5em{\scshape i\kern-0.25em b}\kern-0.8em\TeX}}}

\copyrightyear{2025}
\acmYear{2025}
\setcopyright{acmlicensed}\acmConference[EuroSys '25]{Twentieth European Conference on Computer Systems}{March 30-April 3, 2025}{Rotterdam, Netherlands}
\acmBooktitle{Twentieth European Conference on Computer Systems (EuroSys '25), March 30-April 3, 2025, Rotterdam, Netherlands}
\acmDOI{10.1145/3689031.3696075}
\acmISBN{979-8-4007-1196-1/25/03}

\usepackage{subfigure}
\usepackage{multirow}
\usepackage{booktabs}
\usepackage{blindtext}
\usepackage{makecell}
\usepackage{tabularx}
\usepackage{tikz}
\usepackage{amsmath}
\usepackage{amsthm}
\usepackage{algorithm}
\usepackage[noend]{algorithmic}
\usepackage{mathtools}
\usepackage{multirow,multicol}
\usepackage{multicol}
\usepackage{ifthen}

\usepackage{amssymb}
\usepackage{optidef}
\usepackage{wrapfig}
\usepackage{graphicx}
\usepackage{subcaption}
\usepackage{float}

\newcommand{\sysname}[0]{{HybridFlow}}

\begin{document}

\title{HybridFlow: A Flexible and Efficient RLHF Framework}
\renewcommand{\shorttitle}{HybridFlow: A Flexible and Efficient RLHF Framework}

\author{Guangming Sheng}
\affiliation{%
  \institution{The University of Hong Kong}
  \country{}
  }
\email{gmsheng@connect.hku.hk}

\author{Chi Zhang}
\affiliation{%
  \institution{ByteDance}
  \country{}
  }
\email{zhangchi.usc1992@bytedance.com}

\author{Zilingfeng Ye}
\affiliation{%
  \institution{ByteDance}
  \country{}
  }
\email{yezilingfeng@bytedance.com}

\author{Xibin Wu}
\affiliation{%
  \institution{ByteDance}
  \country{}
  }
\email{wuxibin@bytedance.com}

\author{Wang Zhang}
\affiliation{%
  \institution{ByteDance}
  \country{}
  }
\email{zhangwang.nozomi@bytedance.com}

\author{Ru Zhang}
\affiliation{%
  \institution{ByteDance}
  \country{}
  }
\email{zhangru.1994@bytedance.com}

\author{Yanghua Peng}
\affiliation{%
  \institution{ByteDance}
  \country{}
  }
\email{pengyanghua.yanghua@bytedance.com }

\author{Haibin Lin}
\affiliation{%
  \institution{ByteDance}
  \country{}
  }
\email{haibin.lin@bytedance.com }

\author{Chuan Wu}
\affiliation{%
  \institution{The University of Hong Kong}
  \country{}
}
\email{cwu@cs.hku.hk}

\renewcommand{\shortauthors}{G. Sheng, C. Zhang, Z. Ye, X. Wu, W. Zhang, R. Zhang, P. Yang, H. Lin, C. Wu}




\begin{abstract}
Reinforcement Learning from Human Feedback (RLHF) is widely used in Large Language Model (LLM) alignment.
Traditional RL can be modeled as a %
dataflow, where each node represents computation of a neural network (NN) %
and each edge denotes data dependencies between the NNs. %
RLHF complicates the dataflow by expanding each node into a distributed LLM training or generation program, and each edge into a many-to-many multicast.
Traditional RL frameworks execute the dataflow using a single controller to instruct both intra-node computation and inter-node communication, %
which can be \textit{inefficient} in RLHF due to large control dispatch overhead for distributed intra-node computation. %
Existing RLHF systems %
adopt a multi-controller paradigm, which can be \textit{inflexible} due to nesting distributed computation and data communication. %
We propose \textit{\sysname}, which combines single-controller and multi-controller paradigms in a \underline{\textit{\smash{hybrid}}} manner to enable
\textit{flexible} representation and \textit{efficient} execution of the RLHF data\underline{\textit{\smash{flow}}}.
We carefully design a set of hierarchical APIs that decouple and encapsulate computation and data dependencies %
in the complex RLHF dataflow, allowing efficient operation orchestration to implement RLHF algorithms and flexible mapping of the computation onto various devices. %
We further design a 3D-HybridEngine for efficient actor model resharding between training and generation phases, with zero memory redundancy and significantly reduced communication overhead.
Our experimental results demonstrate 1.53$\times$$\sim$20.57$\times$ throughput improvement when running various RLHF algorithms using \sysname{}, as compared with state-of-the-art baselines. 
\sysname{} source code will be available at \href{https://github.com/volcengine/verl}{https://github.com/volcengine/verl}

\end{abstract}

\maketitle

\section{Introduction} \label{sec:intro}

Large language models (LLMs) such as %
GPT~\cite{gpt3}, Llama~\cite{touvron2023llama} and Claude~\cite{bai2022training} have revolutionized various artificial intelligence (AI) applications, ranging from writing~\cite{openai2023gpt4}, searching~\cite{nakano2021webgpt} to coding~\cite{rozière2023codellama}. 
LLMs are first pre-trained on trillions of tokens from books, websites, etc,. via next-word prediction to accumulate broad knowledge \cite{gpt3}. Next, LLMs are trained on domain-specific datasets via supervised fine-tuning (SFT), to be able to follow human instructions~\cite{gpt3}. Despite the outstanding capabilities of LLMs on natural language tasks after pre-training and SFT, the detrimental and biased contents in the training datasets may still mislead an LLM to generate toxic and undesirable content. Reinforcement Learning from Human Feedback (RLHF) %
is introduced to further align an LLM to human values, for building helpful and harmless AI applications~\cite{bai2022training, ouyang2022training}.

RLHF is built upon %
traditional RL algorithms~\cite{schulman2017proximal, williams1992REINFORCE, akrour2011preference_tradtionrl}, e.g., Proximal Policy Optimization 
(PPO)~\cite{schulman2017proximal} and REINFORCE~\cite{williams1992REINFORCE}.
The widely adopted PPO-based RLHF system typically consists of four LLMs~\cite{bai2022training, ouyang2022training}%
: an \textit{actor}, a \textit{critic}, a \textit{reference policy} network and a \textit{reward model}.
PPO-based RLHF proceeds in iterations, each with three stages: (1) response \textit{generation} using the actor model with a batch of prompts;
(2) \textit{preparation} of 
training data by scoring the generated responses through a single forward pass of the critic, reference policy, and reward models;
(3) \textit{learning} from human preference by updating actor and critic through forward and backward computation.
Other RLHF variants~\cite{daiSafeRLHFSafe2023, li2023remax} follow similar stages but involves different numbers of models and data dependencies among the models.

Traditional RL can be %
modeled as a %
dataflow%
~\cite{liang2021rllib}, which is a directed acyclic graph (DAG): each node in the RL dataflow represents computation of a neural network (e.g., actor or critic network which can be CNN or MLP);
 each edge denotes data dependency between NN computations (e.g., output of the critic is used as input to actor training~\cite{schulman2017proximal}.) %
RLHF dataflow is more complex, with more complicated models involved 
(e.g., LLMs for the actor/critic/reference/reward models),
each running distinct computation,
and more diverse data dependencies among them (i.e., multicast between distributed model partitions). %
Training and generation of an LLM in the RLHF dataflow requires %
distributed computation %
(e.g., using tensor/pipeline/data parallelism)~\cite{shoeybi2019megatron, kwon2023efficient}. %
Therefore, each node in the RLHF dataflow is a complex distributed program, corresponding to distributed computation of the respective LLM. 
Models in different nodes typically use different parallelism strategies as their workloads vary.
The edge represents data resharding, which is often a many-to-many multicast.
Consequently, 
\textit{\textbf{Flexible}} representation and \textit{\textbf{efficient}} execution of the complex and resource intensive RLHF is imperative.

Traditional RL frameworks such as RLLib~\cite{liang2018rllib} and RLLib Flow~\cite{liang2021rllib} utilize a hierarchical single-controller paradigm to run %
RL dataflows. 
A centralized controller %
assigns nodes in the dataflow to different processes and coordinates their execution order. %
Each node process can further spawn more workers to perform computation, again following the single-controller paradigm. 
However, they only provide primitives for data-parallel training and are constrained to neural networks that are at most hundreds of MB in size~\cite{liang2021rllib, liang2018rllib}.
In the RLHF dataflow, %
each node corresponds to %
an LLM with up to billions of operators, computed using some complex parallelism. 
A single-controller paradigm %
is inefficient due to the substantial overhead of dispatching operators to distributed accelerators%
~\cite{barham2022pathways, abadi2016tensorflow}.

Existing RLHF systems adopt a multi-controller paradigm to manage intra-node computation and inter-node data resharding~\cite{hu23openrlhf,NeMoAligner,xiao2023adaptive}.
Each controller independently manages the computation of one device and uses multiple point-to-point operations to coordinate data dependencies between different nodes.
This multi-controller paradigm introduces negligible dispatch overhead when performing LLM computation (detailed in \textsection\ref{sec:motivate_programming_model}).

However, without central control, it is \textit{inflexible} to implement various RLHF dataflow, as modifying a single node to adapt to different data dependencies requires
changing all dependent nodes' implementation, 
 hindering code reuse.

To address these limitations, we propose \textit{\sysname}, a flexible and efficient RLHF %
framework to easily represent and execute diverse RLHF dataflows, %
attaining high throughput.
Our key observation is that utilizing the single-controller paradigm on the inter-node level enables flexible expression of various data dependencies and easy coordination of inter-node data resharding with minimal overhead,
while integrating the multi-controller paradigm within intra-node computation enhances computation efficiency substantially.
We %
advocate a hierarchical hybrid programming model to generate RLHF dataflows.
At the node level, multiple model classes are provided that encapsulate distributed computation (training, inference and generation) of different LLMs in the dataflow into primitive APIs. %
These APIs %
can seamlessly support various parallelism strategies from the existing LLM %
frameworks, including 3D parallelism~\cite{shoeybi2019megatron}, ZeRO~\cite{rajbhandari2020zero}, and PyTorch FSDP~\cite{paszke2019pytorch}%
), and perform distributed computation under the multi-controller paradigm. %
Among the nodes, a set of transfer protocols are designed to hide the complexity of data resharding from users, as coordinated by a single controller.
This programming model abstracts away the complexity of distributed computing, allowing users to implement an RLHF dataflow in a few lines of code and run RLHF through a single process of the single controller.
It also effectively decouples intra-node computation and inter-node data transfer,
allowing %
independent optimization of each model 
without changing the code of other models in the dataflow.

Training and generation of the actor model represent major computation %
in the RLHF dataflow.
We further design a \textit{3D-HybridEngine} to enable efficient execution of training and generation of the actor model, introducing zero memory redundancy and significantly reduced communication overhead during model parameter resharding between the training and generation stages.
Our hybrid programming model also facilitates flexible placement of models onto the same or different sets of GPU devices. %
This allows us to provide an effective algorithm to optimize GPU allocation and placement of the models, with various model sizes and distinct workloads, for any RLHF dataflow.
Our contributions in designing \sysname{} are summarized as follows:

\noindent$\bullet$ We propose a hierarchical hybrid programming model for conveniently building the RLHF dataflow. This programming model enables efficient distributed execution of intra-node computation and flexible inter-node data resharding and transfer, for %
various RLHF algorithms (\textsection\ref{sec:programming_model}).

\noindent$\bullet$ We design a 3D-HybridEngine that executes training and generation of the actor model with high computation efficiency and zero-redundancy transition between the training stage and the generation stage %
(\textsection\ref{sec:hybrid_engine}).

\noindent$\bullet$ We devise an effective mapping algorithm to automatically identify optimized %
GPU allocation and placement of each node (model) in the RLHF dataflow %
(\textsection\ref{sec:auto_mapping}).

\noindent$\bullet$ 
We conduct extensive experiments comparing \sysname{} with state-of-the-art RLHF systems~\cite{hu23openrlhf, yao2023deepspeedchat, NeMoAligner} under various RLHF algorithms, model sizes and %
cluster scales. %
Our evaluation 
demonstrates 1.53$\times$$\sim$20.57$\times$ throughput improvements.

We have open-sourced \sysname{} and believe that \sysname{} can boost future RLHF research and development.

\begin{figure}[t]
    \centering
    \includegraphics[width=\linewidth]{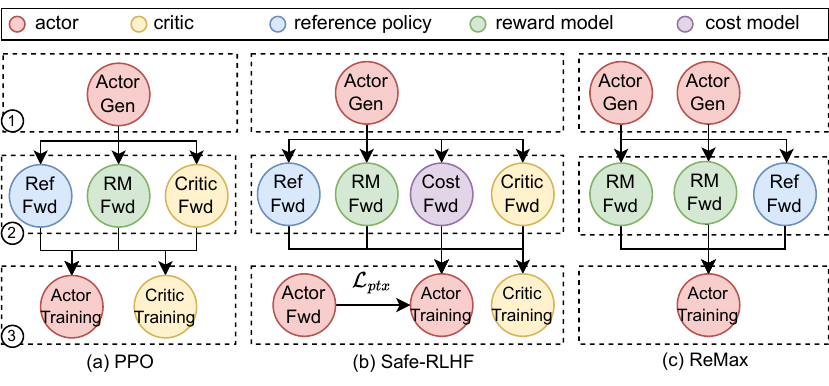}
    \vspace{-6.6mm}
    \caption{Dataflow graph of 3 RLHF algorithms \cite{ouyang2022training, daiSafeRLHFSafe2023, li2023remax}. 
    Stage $\textcircled{1}$, $\textcircled{2}$,  $\textcircled{3}$
    represent Generation, Preparation, and Training, respectively.
    }
    \vspace{-2.4mm}
    \label{fig:rlhf_dataflow}
\end{figure}

\section{Background and Motivation} \label{sec:2_bckground_and_motivation}

\subsection{Reinforcement Learning from Human Feedback} \label{sec:2_1_bckgroud_rlhf}
\noindent\textbf{RLHF Workflow.}
RLHF %
aligns the linguistic space of %
LLMs with human values, using a set of human-ranked candidates of given prompts~\cite{ouyang2022training, bai2022training, daiSafeRLHFSafe2023, shao2024deepseekmath, li2023remax, zheng2023improving, lee2023rlaif}.
An RLHF system typically consists of multiple models, e.g., an actor, a critic, a reference policy, and one or multiple
reward models. The actor and the reference are each pre-trained/fined-tuned LLM (i.e., the LLM that is undergoing RLHF). The critic and reward models can be different LLMs fine-tuned on the human preference dataset, with the language
modeling head replaced by a scalar output head~\cite{ouyang2022training, bai2022training}.
The RLHF workflow can be decomposed into 3 stages (Figure~\ref{fig:rlhf_dataflow}) and we take PPO as an example:

\noindent $\bullet$\textit{Stage 1 (Generation):} The actor produces responses from a batch of prompts using auto-regressive generation.

\noindent $\bullet$\textit{Stage 2 (Preparation): }Using prompts and generated responses, the critic computes their values~\cite{schulman2017proximal, schulman2015trust}, the reference policy computes their reference log probabilities, and the reward model computes their rewards~\cite{ouyang2022training, bai2022training}, all via a single pass of forward computation of the respective model. %

\noindent $\bullet$\textit{Stage 3 (Learning/Training): }The actor and the critic are updated via Adam~\cite{kingma2017adam}, 
using the batch of data produced by previous stages and the loss function~\cite{ouyang2022training}.

Other RLHF algorithms  %
largely follow the 3-stage workflow as well (Figure~\ref{fig:rlhf_dataflow}(b)(c)).
Safe-RLHF~\cite{daiSafeRLHFSafe2023} introduces an auxiliary pretrain loss following PPO-ptx~\cite{ouyang2022training} %
and includes an additional cost model to fit human preferences and safety labels simultaneously. 
ReMax~\cite{li2023remax} %
requires an additional generation pass for variance reduction and eliminates the critic model in the dataflow.
Researchers are actively exploring novel RLHF algorithms~\cite{shao2024deepseekmath, lee2023rlaif, zheng2023improving} and integrating traditional RL methods into RLHF domains~\cite{kaufmann2023survey}. These variances necessitate a flexible representation of the RLHF dataflow graph to accommodate diverse algorithmic requirements.

\begin{figure}[!t]
    \includegraphics[width=\linewidth]{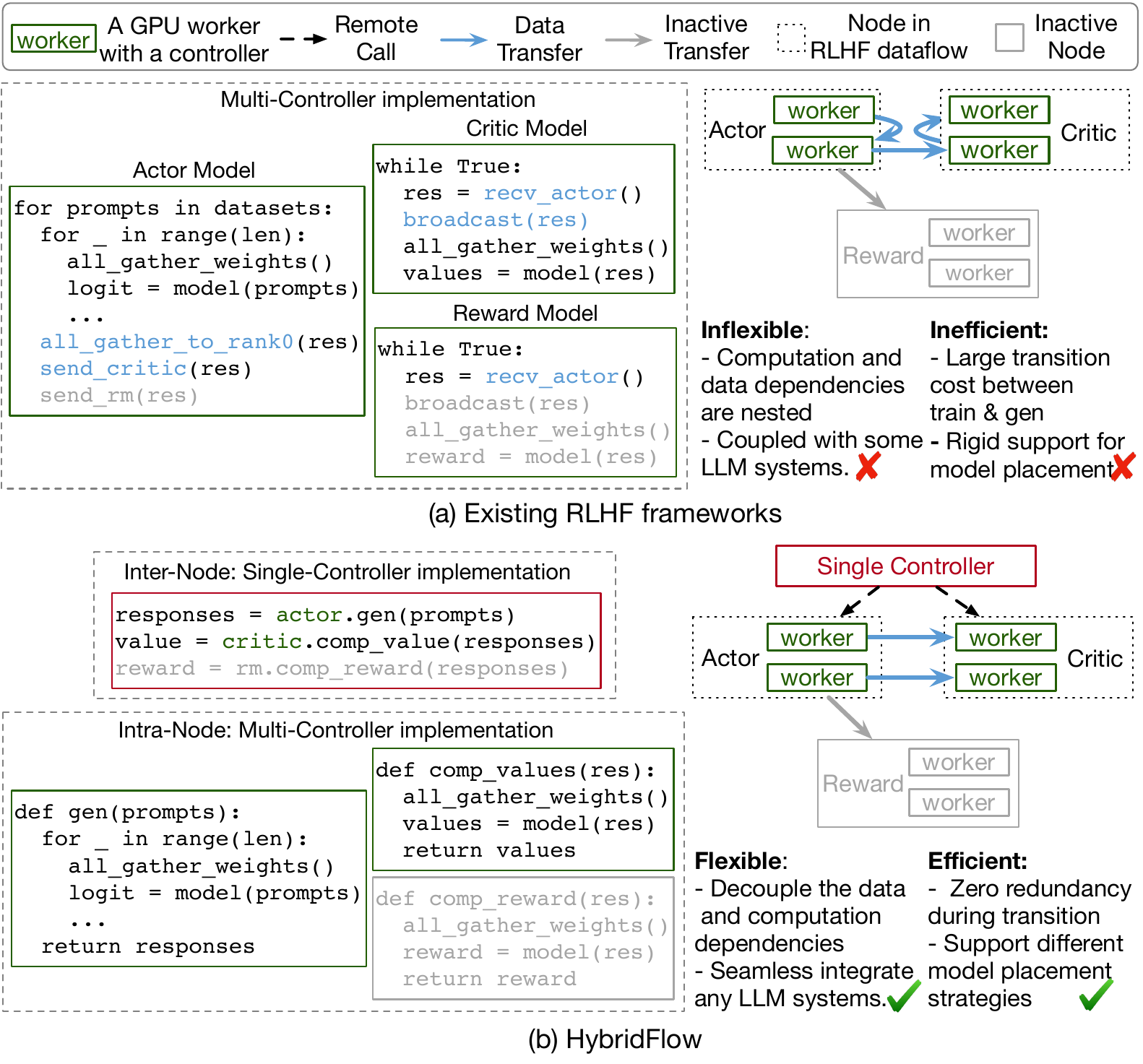}
    \vspace{-5mm}
    \caption{
    Programming model used in RLHF systems. (a) Existing RLHF systems adopt the \textit{multi-controller} paradigm.
    (b) \sysname{} utilizes a hybrid programming model: the \textit{single-controller} coordinates models; each model uses \textit{multi-controller} paradigm in distributed computation. Inactive node in grey represents operation not executed at this time. 
    }
    \label{fig:program_model_compare}
    \vspace{-3mm}
\end{figure}

\noindent\textbf{Parallelism Strategies.}
LLMs are trained and served with data, pipeline, and tensor parallelism~\cite{narayanan2021efficient,jiang2024megascale, kwon2023efficient}. 
With data parallelism (DP), the input data is split into multiple subsets; each subset is processed %
by a separate %
device (e.g., a GPU)~\cite{sergeev2018horovod}. 
ZeRO~\cite{rajbhandari2020zero} is a memory-optimized solution for DP training, progressively sharding optimizer states, gradients, and model parameters across GPUs.
Pipeline parallelism (PP)~\cite{huang2019gpipe, narayanan2019pipedream} and tensor parallelism (TP)~\cite{shoeybi2019megatron}
distribute model parameters, gradients and optimizer states across multiple GPUs. 
Modern distributed training frameworks like Megatron-LM~\cite{shoeybi2019megatron} and MegaScale~\cite{jiang2024megascale} utilize \textit{3D parallelism} or \textit{PTD parallelism} %
~\cite{narayanan2021efficient}, where P, T, D stand for PP, TP, DP, respectively. %
In 3D parallelism, PP size represents the number of pipeline stages in model training, TP size refers to the number of shards that a tensor is partitioned into, and DP size is the number of model replicas.
LLM serving systems
employ 3D parallelism similar to training while only model parameters and KVCache are sharded~\cite{kwon2023efficient, nvidiaTensorRTLLM, holmes2024deepspeed}.

LLM models in the RLHF dataflow may %
perform distinct computations, including training (one forward pass, one backward pass and model update), inference (one forward pass) and generation (auto-regressive generation with multiple forward passes). In particular, training and generation are performed on the actor model, training and inference on the critic, and inference on reference policy and reward models.
Distinct parallel strategies can be applied to different models for varied computations to achieve optimal throughput. %

\subsection{Programming Model for Distributed ML} \label{sec:motivate_programming_model}

\noindent \textbf{Single-Controller.} %
It employs a centralized controller to manage the overall execution flow of the distributed program. With centralized control logic, users can build core functionalities of the dataflow as a single process (Figure~\ref{fig:program_model_compare}(b)), while the controller automatically generates distributed workers to carry out the computation.
With a global view of the hardware and dataflow graph, the single-controller paradigm allows flexible and optimized %
resource mapping and execution order coordination among dataflow tasks. 
However, coordination messages are passed from the controller to all workers, %
incurring significant dispatch overhead when executing expansive dataflow graphs on large clusters~\cite{abadi2016tensorflow, barham2022pathways}.

\noindent \textbf{Multi-Controller.} %
Each device (aka worker) has its own controller. %
State-of-the-art distributed LLM training and serving systems adopt the multi-controller paradigm, due to its scalability and low dispatch overhead (control messaging largely passed from CPU to GPU over fast PCIe links) %
~\cite{shoeybi2019megatron, jiang2024megascale, rasley2020deepspeed, kwon2023efficient}.
As shown in the example that employs multi-controller RLHF implementation in Figure~\ref{fig:program_model_compare}(a), %
a separate program is run for each model, and all workers of one model execute the same program.
Each %
worker only possesses a local view of the system state and requires %
point-to-point communication between two models %
(blue code and arrows) to coordinate model execution order.
To implement an RLHF workflow in the multi-controller architecture, a user must intricately integrate the code for collective communication, computation, and point-to-point data transfer in the program run at each device.
This leads to deeply nested code of %
computation and data transfer, %
challenging to develop, maintain, and optimize. 
In Figure~\ref{fig:program_model_compare}(a), each model performs local computation and all\_gather operations (black code), while the actor model must explicitly manage send operations to the critic and reward models, and the latter must correspondingly implement receive operations at precise points in their program.

\vspace{-2mm}
\subsection{RLHF Characteristics} \label{sec:2_3_rlhf_characterisitc}

\noindent\textbf{Heterogeneous model workloads.} 
The actor, critic, reference and reward models in RLHF may %
execute training, inference or generation at different stages, with different %
memory footprint and computation demand.
For reference policy and reward models,  %
only their model parameters need to be stored in GPU memory, as they
perform only the forward pass computation.
For the actor and the critic, their model parameters, gradients, and optimizer states must be stored as they undergo model training. %
Moreover, a small actor model (e.g., a 7B pre-trained/fine-tuned LLM) can be paired with %
larger critic and reward models (e.g., 70B LLMs) in RLHF for better alignment~\cite{bai2022training}.
Given such heterogeneity, different parallelism strategies and tailored %
optimizations are needed for running each model during RLHF.

\noindent\textbf{Unbalanced computation between actor training and generation.} \label{sec:hybrid_motivation}
In the RLHF dataflow, training and generation of the actor model are represented by two nodes (Figure~\ref{fig:rlhf_dataflow}), which often render majority of the workload in each RLHF iteration ({e.g., 58.9\% of total RLHF time with \sysname{}}).
Actor training is computation bound~\cite{geoffrey2021habitat_computebound}, %
often requiring a larger model-parallel (MP) size (i.e., the number of partitions the model is partitioned into) and distributing the workload to more GPUs, e.g., {8 partitions of a 7B model on 8 GPUs.}
Using the same parallelism strategy (e.g., the same MP size) for generation can lead to underutilization of GPU computation resources due to its memory-bound nature~\cite{kwon2023efficient}. 
Previous studies show that combining %
a larger DP size with a smaller MP size (hybrid data and model parallelism), e.g., 
{partition a 7B model into two and replicate it four times on 8 GPUs},
can improve the generation throughput~\cite{li2023alpaserve, zhongDistServeDisaggregatingPrefill2024}. 
Although using different parallelism strategies for actor training and generation may optimize throughput in both stages, resharding the actor model weights at runtime between the two stages can incur significant communication and memory overhead. %
For example, aligning a 70B actor model requires transferring 140GB of model weights from training to generation per RLHF iteration, taking up to 36.4\% of an iteration time when the two stages are on different devices~\cite{hu23openrlhf}.

\begin{figure}[t]
    \includegraphics[width=\linewidth]{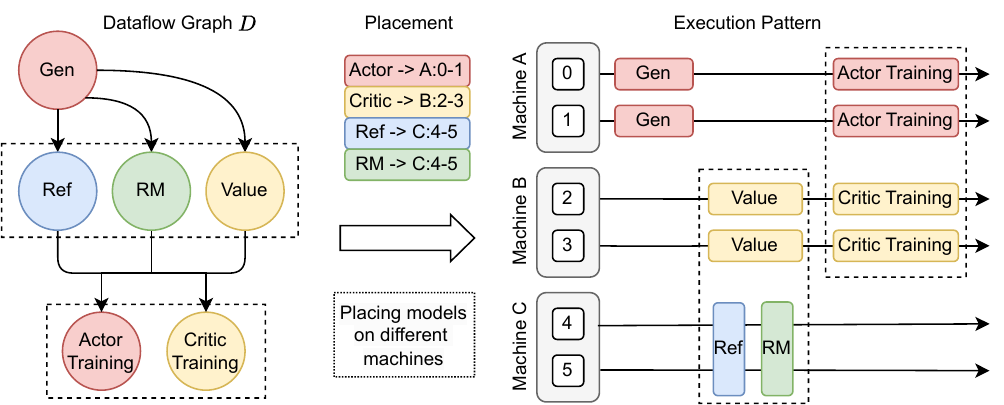}
    \vspace{-5mm}
    \caption{Dataflow execution given a model placement plan. Blocks with numbers represent GPUs. In dashed boxes, the models are placed on different sets of devices and can be concurrently computed. Reference model (blue) and reward model (green) are colocated on the same set of GPUs and executed sequentially.}
    \vspace{-3mm}
    \label{fig:placement}
\end{figure}

\begin{table*}[t]
    \centering   
    \caption{Comparison of RLHF frameworks. Figures illustrate execution of one PPO iteration. Numbers 1-6 %
    represent response generation, reward model inference, reference model inference, critic inference, actor training, and critic training, respectively. %
    }
    \vspace{-3mm}
    \resizebox{\linewidth}{!}{%
    
    \begin{tabular}{c|c|c|c|c}
        \toprule
        RLHF system& DeepSpeed-Chat & OpenRLHF & NeMo-Aligner  &\textbf{HybridFlow} \\
        \midrule
        \makecell{Parallelism\\} & \makecell{Training: ZeRO \\ Generation:TP} & \makecell{Training: ZeRO \\ Generation:TP}  & \makecell{3D Parallelism for both\\ training and generation} & 
        \makecell{\textbf{Training: 3D, ZeRO, FSDP} \\ \textbf{Generation: 3D Parallelism}} 
        \\
        \midrule
        \makecell{Actor weights \\ in training \& generation} & \makecell{Model resharding\\ from ZeRO to TP} & \makecell{Using two copies of actor \\ weights for the two stages}  & \makecell{Using identical model partition \\ in two stages (shared weights)} &\makecell{\textbf{Zero-redundancy} \\ \textbf{model resharding} }\\
        \midrule
        \makecell{Model\\Placement} & \makecell{Colocate all models \\on the same set of devices} & \makecell{Each model placed \\on separate devices} & \makecell{Actor/Ref colocated on some GPUs\\Critic/RM colocated on other GPUs} &\textbf{\makecell{Support various \\ model placement}} \\
        \midrule
        \makecell{Execution\\Pattern\\
        \begin{minipage}{2cm}
            \includegraphics[width=\linewidth]{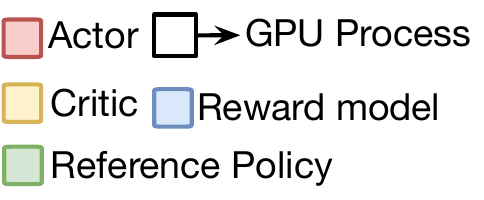}
        \end{minipage}}
        &
        \begin{minipage}{2.5cm}
            \includegraphics[width=\linewidth]{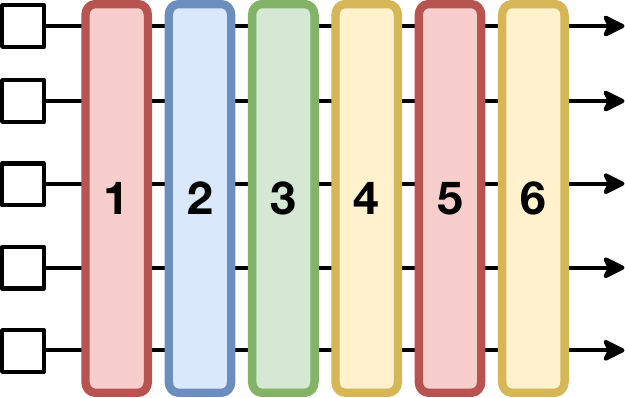}
        \end{minipage}
        &
        \begin{minipage}{2.3cm}
            \includegraphics[width=\linewidth]{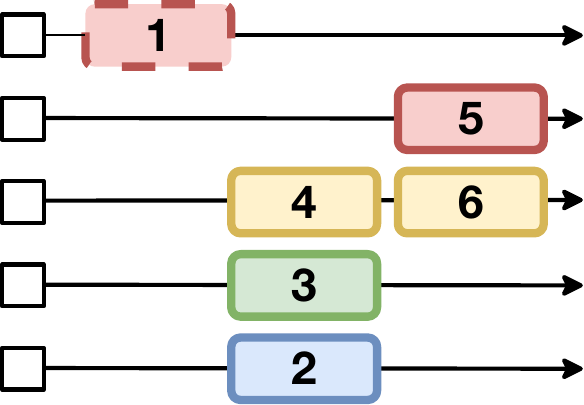}
        \end{minipage}
        &
        \begin{minipage}{2.3cm}
            \includegraphics[width=\linewidth]{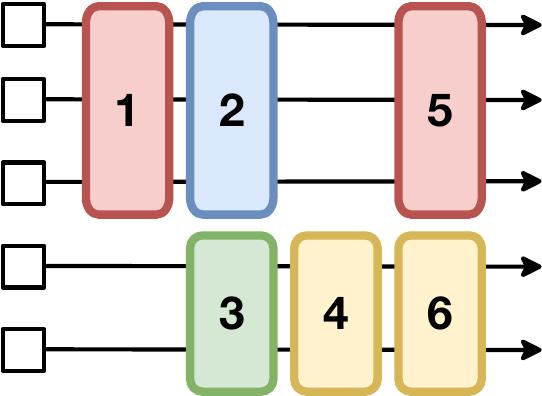}
        \end{minipage}
        &
        \textbf{\makecell{Support various \\execution patterns}}
        \\
        \bottomrule
    \end{tabular}%
    }
    \label{tab:table_with_images}
    \vspace{-3mm}
\end{table*}

\noindent\textbf{Diverse model placement requirements.}
Strategic device placement of models in the RLHF dataflow is necessary, %
according to computation workloads and data dependencies of the models. Figure~\ref{fig:placement} gives an example model placement plan and the corresponding RLHF execution flow. %
Models placed on different sets of devices can be executed in parallel if no data dependencies exist. Models placed on the same set of GPUs, referred to as {\em colocated models}, share the GPU memory and are executed sequentially in a time-sharing manner, as out-of-memory (OOM) error may easily happen if colocated LLMs execute concurrently.  

We observe a compromise: placing models on different devices permits parallel processing but may inevitably lead to some GPU idle time, given staged model execution in RLHF.
In Figure~\ref{fig:placement}, actor and critic are placed separately, performing training in parallel, but incurring 1/3 of their GPU time being idle, during other RLHF stages. Supporting various placement strategies and maximizing device utilization %
are crucial for optimizing RLHF performance at any model size and cluster scale.

\vspace{-1mm}
\subsection{Limitations of existing RLHF systems}
\vspace{-1mm}
\noindent\textbf{Inflexible support for various RLHF dataflow graphs.}
Existing RLHF systems %
adopt the multi-controller paradigm for dataflow implementation~\cite{yao2023deepspeedchat, hu23openrlhf, NeMoAligner, xiao2023adaptive}. %
{To implement various RLHF algorithms,}
a user must navigate and manage code %
that mixes collective communication, model computation (potentially using various distributed training/serving frameworks), and point-to-point data transfer. %
This %
code structure lacks %
modularity/function encapsulation, making the RLHF systems tightly coupled with specific LLM training and serving frameworks.
Consequently, a user needs to implement and optimize different RLHF dataflows case-by-case~\cite{liang2021rllib}, hindering code reuse and increasing the risk of making mistakes. Existing RLHF frameworks only support the PPO algorithm.
In addition, limited parallel strategies are supported due to implementation complexity. For example, to incorporate 3D parallelism for LLM training and generation
in DeepSpeed-Chat~\cite{yao2023deepspeedchat}, one may have to re-implement the whole system due to the mixed code structure.

\noindent\textbf{Inefficient RLHF execution.} %
Table~\ref{tab:table_with_images} summarizes parallelism strategies, model placement, and execution patterns adopted by the existing RLHF systems. %
DeepSpeed-Chat~\cite{yao2023deepspeedchat} and OpenRLHF~\cite{hu23openrlhf} adopt ZeRO-3 for actor training and TP for actor generation.
OpenRLHF uses different copies of the actor model on different devices for training and generation,
incurring redundant memory usage and frequent weight synchronization among devices.
DeepSpeed-Chat maintains the same copy of actor model on the same set of devices for training and generation, %
and reshards model weights between training and generation (due to different parallelisms used in the two stages), %
which may still incur substantial {memory and communication} overhead 
for large models (detailed in \textsection\ref{sec:hybrid_comm_mem}). 
NeMo-Aligner~\cite{NeMoAligner} uses the same 3D parallelism configurations in actor training and generation, %
experiencing low generation throughput  (\textsection\ref{sec:exp_benefit_hybrid_engine}).

Existing RLHF frameworks are limited to one model placement plan and hence one RLHF execution pattern, as shown in Table~\ref{tab:table_with_images}. Implementing a different placement is difficult, requiring changing the inner logic of 
{model initialization and inter-node data transfer as highlighted in blue in Figure~\ref{fig:program_model_compare}.}
OpenRLHF and NeMo-Aligner allow concurrent model computation in the preparation and learning stages; in the generation stage, models except the actor are idle, wasting the GPUs they occupy. 
DeepSpeed-Chat %
colocates all models on the same set of devices, 
and each device runs each model sequentially according to the RLHF dataflow. With unbalanced workloads among the models, such a placement can be inefficient in resource utilization
(evaluated in \textsection\ref{sec:exp_placement}).

\vspace{-2mm}
\subsection{Design Considerations}
To tackle limitations of existing systems,
the key question is -
\textbf{How to design a flexible and efficient programming model to implement RLHF dataflow?}
A single-controller design is particularly advantageous at the inter-node level due to its flexibility in coordinating data transfer, execution order, and resource virtualization among distributed computation of different models~\cite{barham2022pathways, moritz2018ray}. 
The RLHF dataflow graph  %
typically consists of only a few nodes. %
Dispatching control messages to different nodes from the single-controller %
incurs negligible overhead as compared to distributed computation required for nodes (models) in the dataflow.
The multi-controller paradigm, known for its low latency in dispatching operators to accelerators~\cite{darema2001spmd}, can be leveraged in distributed computation of each model.
With these insights, we propose a hierarchical hybrid programming model for RLHF dataflow implementation.
Our key design principle is to combine single-controller and multi-controller paradigms in a hybrid manner. %
This design ensures flexible expression and efficient execution of RLHF dataflow, maintaining low control overhead at both inter-node and intra-node levels.
As shown in Figure~\ref{fig:program_model_compare}(b), this paradigm decouples intra-node distributed computation and inter-node data transfer, allowing {each model}
to focus solely on local computation without managing inter-node communication. %

\vspace{-2mm}
\section{\sysname{} Overview} \label{sec:design}

\begin{figure}[t]
    \includegraphics[width=\linewidth]{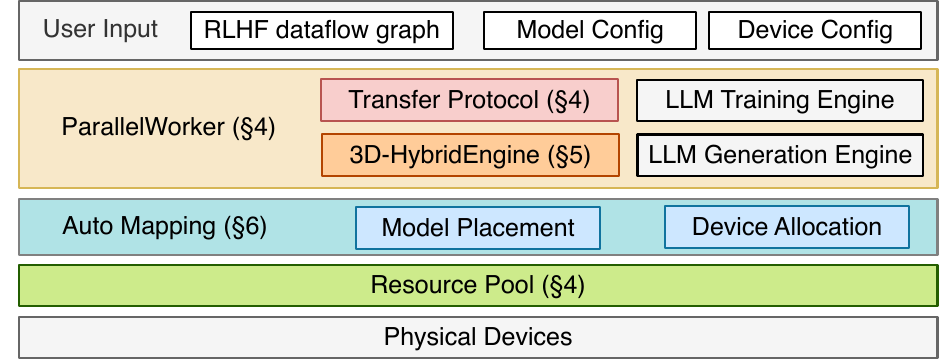}
    \vspace{-5mm}
    \caption{Architecture of HybridFlow. 
    }
    \vspace{-6mm}
    \label{fig:architecture}
\end{figure}

Figure~\ref{fig:architecture} depicts the architecture of \sysname{}, which consists of three major components: 
\textit{Hybrid Programming Model}, \textit{3D-HybridEngine} and \textit{Auto-Mapping algorithm}.
The hybrid programming model includes a set of hierarchical APIs to enable flexible expression of the RLHF dataflow and efficient computation of models in the dataflow (\textsection\ref{sec:programming_model}). The 3D-HybridEngine is particularly designed for efficient training and generation of the actor model, allowing different 3D parallel configurations 
in the two stages and %
enabling zero memory redundancy and minimized communication overhead during the transition between two stages (\textsection\ref{sec:hybrid_engine}).
The auto- mapping algorithm determines optimized device %
placement of each model to maximize the throughput of RLHF (\textsection\ref{sec:auto_mapping}).

The workflow of our RLHF system goes as follows. A user provides the following inputs to start the RLHF system: (i) model specifications, including
{the architecture and size}
of the actor/critic/reference policy/reward models in the RLHF dataflow;
(ii) device placement of the models in the dataflow, as obtained by running the auto-mapping algorithm under given GPU cluster configurations; (iii) parallelism strategy for running each model in each stage, e.g., a tuple of (p, t, d) for 3D parallelism, where p, t, d represent PP size, TP size and DP size, respectively. %
The single controller program takes these inputs to initialize models in the RLHF dataflow and virtualized resource pool, dispatches operations/models to devices according to the placement plan, and invokes functions run by the multiple controllers on devices to carry out distributed computation of each model.

The multi-controller program implements the ParallelWorker class: it %
constructs parallel groups of each model among allocated devices %
according to its parallelism strategies, 
invokes the 3D-HybridEngine for actor training and generation, and can be integrated seamlessly with existing LLM engines~\cite{shoeybi2019megatron, rasley2020deepspeed, paszke2019pytorch, kwon2023efficient} for training, inference and generation of other models.
The transfer protocols are coordinated by the single controller program to support resharding of data (including prompts, responses, and other model outputs in RLHF) between models with distinct parallelism strategies. The data resharding of the actor between training and generation is handled by 3D-HybridEngine.

\vspace{-2mm}
\section{Hybrid Programming Model}  \label{sec:programming_model}

\subsection{Hierarchical APIs}

\begin{figure}[t]
    \includegraphics[width=\linewidth]{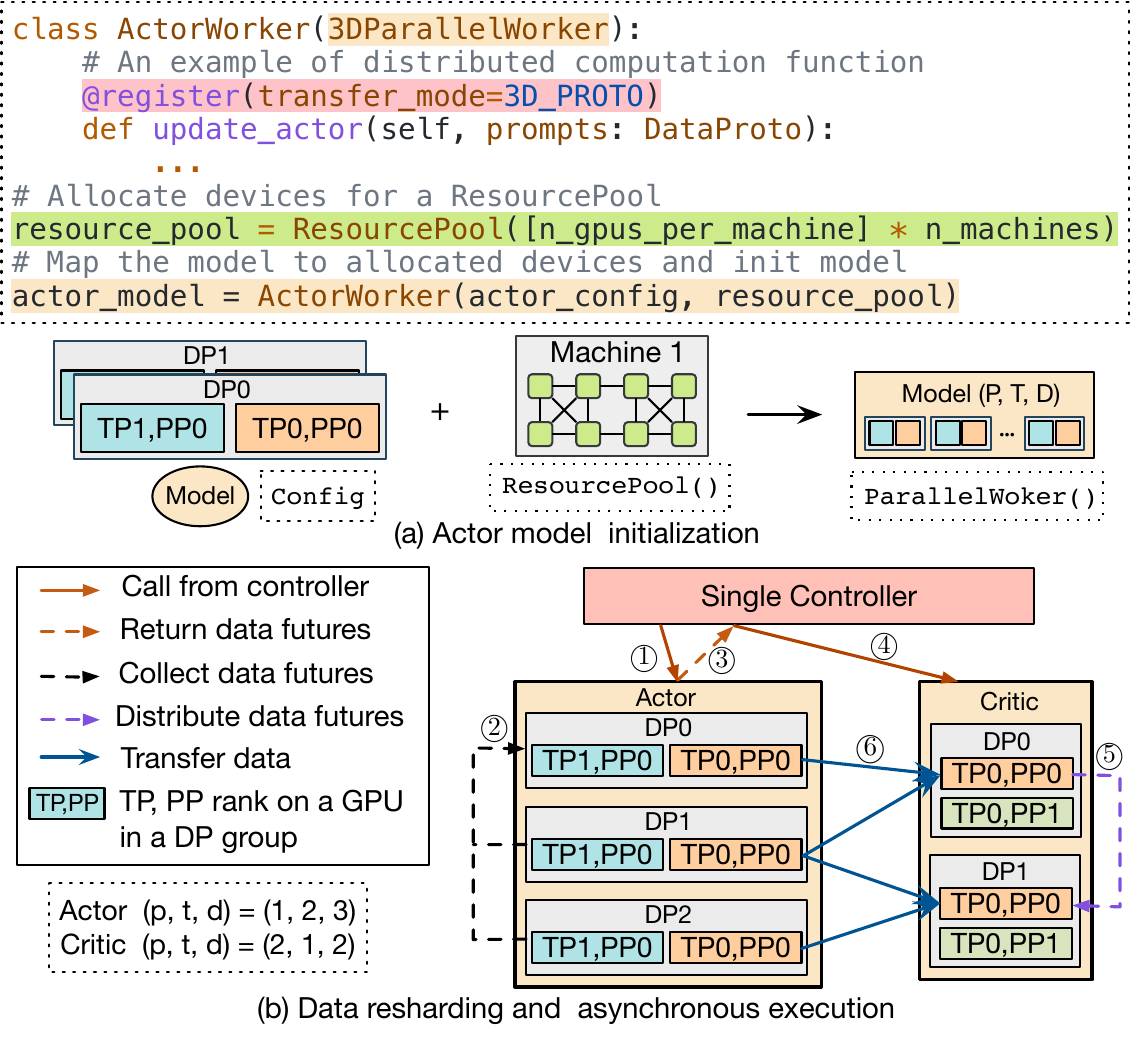}
    \vspace{-8mm}
    \caption{An illustration of hierarchical APIs. (a) Model with 3D parallel configuration, resource allocation, and \texttt{3DParallelWorker} initialization. (b) Asynchronous data resharding between two models with collect and distribute functions in \texttt{3D\_PROTO}.
    }
    \label{fig:api_figure}
    \vspace{-5mm}
\end{figure}

\noindent\textbf{Intra-node: encapsulating distributed program.}
For distributed computation of each model
in different RLHF stages, we provide a base class, \verb|3DParallelWorker|. Given allocated devices, it facilitates distributed model weight initialization and establishes 3D parallel groups for each model. A parallel group includes a set of GPUs to host a specific parallel dimension of the model, e.g., different tensor shards in TP and different model replicas in DP. Figure~\ref{fig:api_figure}(a) illustrates initialization of the actor model with our APIs, while initialization of other models is similar.

Inheriting from the \verb|3DParallelWorker| class, several model classes, for actor, critic, reference, and reward model, respectively, are provided.
Each of these model classes encapsulates APIs to implement the model's distributed forward and backward computation, auto-regressive generation, and optimizer updates, decoupling the distributed computation code with data dependencies with other models.
These APIs can be easily implemented by reusing the computation scripts from existing LLM systems. For example, the computation involved in \verb|update_actor| function of \verb|ActorWorker| (the class for the actor model) is similar to the pre-training scripts in Megatron-LM~\cite{shoeybi2019megatron}. 
A model class encapsulates %
fundamental operations %
for implementing various RLHF algorithms, 
e.g., \verb|generate_sequences| in the actor model class for generating responses based on the prompts 
and \verb|compute_reward| in the reward model class for evaluating responses through a forward pass.
({More APIs are detailed in 
Appendix~\ref{appendix:primitive_apis}}).

Besides base class \verb|3DParallelWorker| that implements 3D parallelism, we further provide base classes for PyTorch FSDP (\verb|FSDPWorker|) and ZeRO (\verb|ZeROWorker|), and the corresponding model classes inheriting each base class, to support different parallelism strategies in model computation. ParallelWorker in Figure~\ref{fig:architecture} denotes one of these base classes.

\vspace{0.5mm}
\noindent\textbf{Inter-node: unifying data resharding implementation between models.} 
Many-to-many multicast is involved for data transfer between models employing different parallelism strategies on different devices. %
We unify this data transfer implementation
by associating each operation in each model class with a transfer protocol, using \verb|@register|. 
Each transfer protocol consists of a collect function and a distribute function, to aggregate output data 
 and distribute input data 
according to the parallelism strategy of each model. In the example in~\autoref{fig:api_figure}(a), \texttt{update\_actor} operation is 
{registered to}
transfer protocol \texttt{3D\_PROTO}, as 3D parallelism is used for actor training.
In \texttt{3D\_PROTO}, the collect function gathers all the {output data of corresponding model function (e.g., the loss scalar return from the \texttt{update\_actor})}
in each DP group to the single controller, and the distribute function distributes the input data {to the registered function (e.g., advantages for the \texttt{update\_actor})}
to each DP group. %
Data resharding is enabled using the source model's output collect function and the destination model's input distribute function. 
Figure~\ref{fig:api_figure}(b) illustrates data resharding between the actor (generation) and the critic (inference), where computation of the models adopts different 3D parallelism strategies.
The single controller gathers data futures using the collect function in \verb|3D_PROTO| of actor (steps $\textcircled{1}$-$\textcircled{3}$)  and sends it to 
critic (step $\textcircled{4}$); critic distributes the received data futures to each DP group using the distribute function in its \verb|3D_PROTO| (step $\textcircled{5}$).
Then %
remote data is retrieved from actor to critic, with each of critic's GPUs only fetching the required local batch of the actor's output data according to its DP rank (step $\textcircled{6}$). The actual data transfer only occurs between GPUs, avoiding any central bottleneck.

We provide 8 transfer protocols, including \verb|3D_PROTO|, \verb|DP| \verb|_PROTO|, \verb|ONE_TO_ALL|, etc., that cover %
most data resharding scenarios 
({detailed in Appendix~\ref{appendix:transfer_protocols}}).
A user can further extend the transfer protocols %
through implementing customized collect and distribute functions.

\noindent\textbf{Facilitating flexible model placement.}
We provide a \verb|ResourcePool| class that virtualizes a set of GPU devices. 
When applying a \verb|ResourcePool| instance to a model class (Figure~\ref{fig:api_figure}(a)), distributed computation of the model will be mapped to the devices.
Models utilizing the same \verb|ResourcePool| instance are colocated on the same set of GPUs; %
models are placed on different sets of GPUs when different \verb|Resource| \verb|Pool| instances are applied in their model classes.
We assume no overlap between different \verb|ResourcePool| instances.

\begin{figure}[t]
    \includegraphics[width=\linewidth]{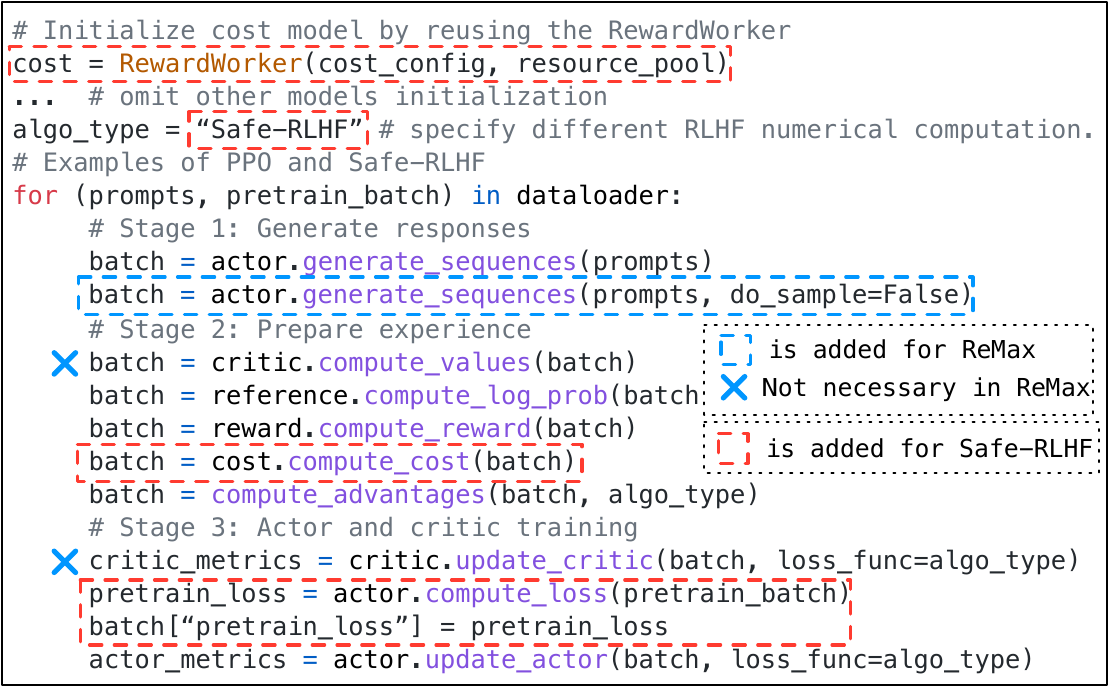}
    \vspace{-7mm}
    \caption{Implementation of PPO~\cite{ouyang2022training}, ReMax~\cite{li2023remax}, and Safe-RLHF~\cite{daiSafeRLHFSafe2023}. Users can adapt to different RLHF algorithms by simply adding or deleting a few lines of code.}
    \label{fig:api_code}
    \vspace{-5mm}
\end{figure}

\noindent\textbf{Asynchronous dataflow execution.}
When models are placed on separate sets of devices, their execution is triggered automatically %
as soon as their inputs become available~\cite {moritz2018ray}.
In Figure~\ref{fig:api_figure}(b), the data future from actor is immediately returned after the controller's call (steps $\textcircled{1}$-$\textcircled{3}$); the controller then initiates a new call to critic and distributes the futures following the %
transfer protocol (steps $\textcircled{4}$-$\textcircled{5}$). 
When some models are placed on the same set of devices, they are executed sequentially based on the calling order. %
With our programming model, \sysname{} is flexible in supporting diverse distributed execution patterns %
without any code change of the RLHF algorithm (Figure~\ref{fig:api_code}).

\subsection{%
Implementation of different RLHF algorithms} \label{sec:examples_of_RLHF_dataflow}

Our APIs %
enable streamlined development of various RLHF algorithms (dataflows). 
Users can implement an RLHF algorithm in a few lines of code as a single process program to run on the single controller,
that involves a sequence of primitive API calls to invoke distributed computation of models. 
Examples of PPO, ReMax, and Safe-RLHF are given in Figure~\ref{fig:api_code}. PPO can be implemented in just 8 lines by invoking model operations
including \texttt{compute\_values} and \texttt{generate\_sequences}, which are executed under the multi-controller paradigm on multiple GPUs.
To adapt to Safe-RLHF which integrates an additional cost model to evaluate safety preferences and %
the pre-taining loss for actor, only 5 more lines of code are added on top of PPO implementation.
{To adapt to ReMax, one additional call to actor generation is needed, and the critic-related code can be removed.}

\noindent \textbf{Achieving flexible.}
This flexibility of extension is crucial for researchers to explore different RLHF algorithms: they can reuse distributed computation encapsulated in each model class and 
simply adjust the code for numerical computations according to specific algorithms, such as GAE~\cite{schulman2018highdimensional} and KL divergence in \verb|compute_advantage| and loss functions of actor and critic.
The streamlined development can be attributed to %
the hybrid programming model.
Our modular API design simplifies development, facilitates extensive code reuse, %
and enables directly incorporating the codebase of existing LLM training/serving frameworks. 
It also decouples model computation and data transfer among models. 
Any change in the distributed frameworks does not affect the code of the RLHF algorithm (Figure~\ref{fig:api_code}), enabling individualized optimization for each model's execution (\textsection\ref{sec:hybrid_engine}).
Flexible placement of models with diverse workloads is supported, %
enabling optimized mapping of RLHF dataflow onto various devices (\textsection\ref{sec:auto_mapping}).

\section{3D-HybridEngine} \label{sec:hybrid_engine}
\begin{figure}[t]
    \centering
    \includegraphics[width=\linewidth]{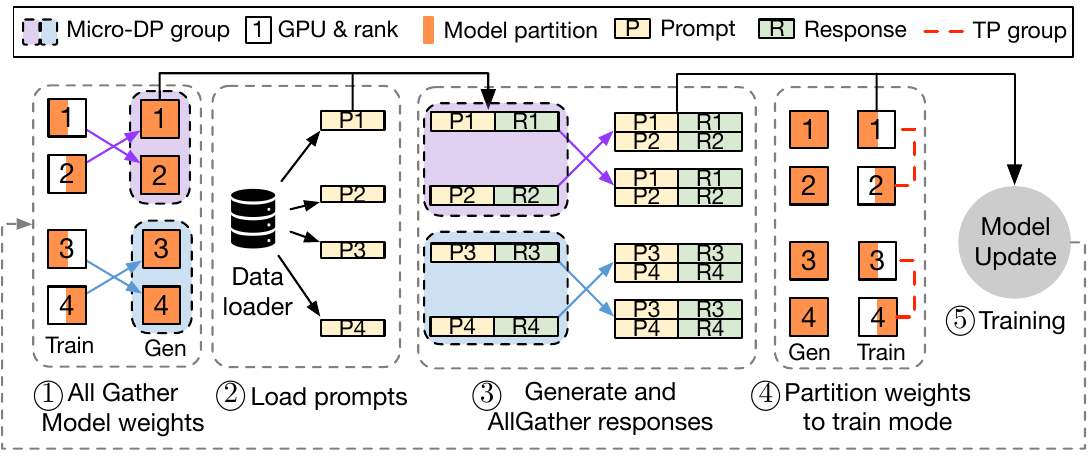}
    \vspace{-5mm}
    \caption{3D-HybridEngine workflow in one RLHF iteration. 4 GPUs are used for actor training and generation. 1-2-2 ($p$-$t$-$d$) parallel groups are used in training and 1-1-2-2 ($p_g$-$t_g$-$d_g$-$d$) parallel groups are used in generation.
    }
    \label{fig:hybrid_one_iter}
    \vspace{-2mm}
\end{figure}

We design the \textit{3D-HybridEngine} %
to support efficient training and generation of the actor model, %
targeting significant RLHF throughput improvement.

\subsection{Parallel Groups}%

To eliminate redundant actor model copies, we advocate deploying actor training and generation stages %
on the same set of devices, $N_a$ GPUs allocated to the actor, and execute them sequentially on the same copy of actor model weights. Nonetheless, actor training and generation may well adopt different 3D parallelism strategies, i.e., the generation stage typically requires smaller TP and PP sizes but a larger DP size, than the training stage (\textsection\ref{sec:hybrid_motivation}). 3D-HybridEngine enables efficient model parameter resharding between actor training and generation across the same set of devices in this context. 

Let $p$-$t$-$d$ denote 3D parallel groups constructed for %
actor training, corresponding to the set of GPUs to host $p$ pipeline stages, $t$ tensor shards, and $d$ model replicas~\cite{narayanan2021efficient}. 3D-HybridEngine builds different parallel groups for actor training and generation, according to their different 3D parallelism strategies, respectively. We use $p_g$, $t_g$, and $d_g$ to denote the size of generation pipeline parallel group, generation tensor parallel group, and micro data parallel group, respectively, 
in the generation stage.  
$d_{g}$ indicates the ratio %
of model replica number in generation over that in training, i.e., each DP replica in training becomes $d_{g}$ micro DP replicas, to process $d_{g}$ microbatches of prompts and responses. We have $N_a$=$p$$\times$$t$$\times$$d$=$p_g$$\times$$t_g$$\times$$d_g$$\times$$d$ such that $d_{g} = \frac{pt}{p_{g}t_{g}}$. %
The micro DP groups are employed exclusively in actor generation stage to render a larger DP size for full device utilization.
The generation parallel groups are denoted by $p_g$-$t_g$-$d_g$-$d$. %

\subsection{3D-HybridEngine Workflow}
Between actor training in iteration $i$ of RLHF and actor generation in iteration $i+1$, the actor model parameters need to be resharded and prompts data to be distributed,
following the parallel group configurations in the two stages.
In iteration $i+1$ of RLHF, %
3D-HybridEngine gathers the actor model parameters updated in iteration $i$ (step $\textcircled{1}$ in Figure~\ref{fig:hybrid_one_iter}), for generation within each micro DP group. Then, the batch of prompts are loaded to each model replica (step $\textcircled{2}$), which generates responses (Generation stage of RLHF). 
Following this, 3D-HybridEngine performs an all-gather operation on the generation results within each micro DP group (step $\textcircled{3}$), and re-partitions %
model parameters according to the 3D parallelism for actor training %
(step $\textcircled{4}$). 
With model weights, prompts and responses correctly re-distributed, %
the loss of the actor model is computed and actor model weights are updated following the RLHF algorithm %
(step $\textcircled{5}$) - actor training stage of iteration $i+1$.

\begin{figure}[t]
    \centering
    \includegraphics[width=\linewidth]{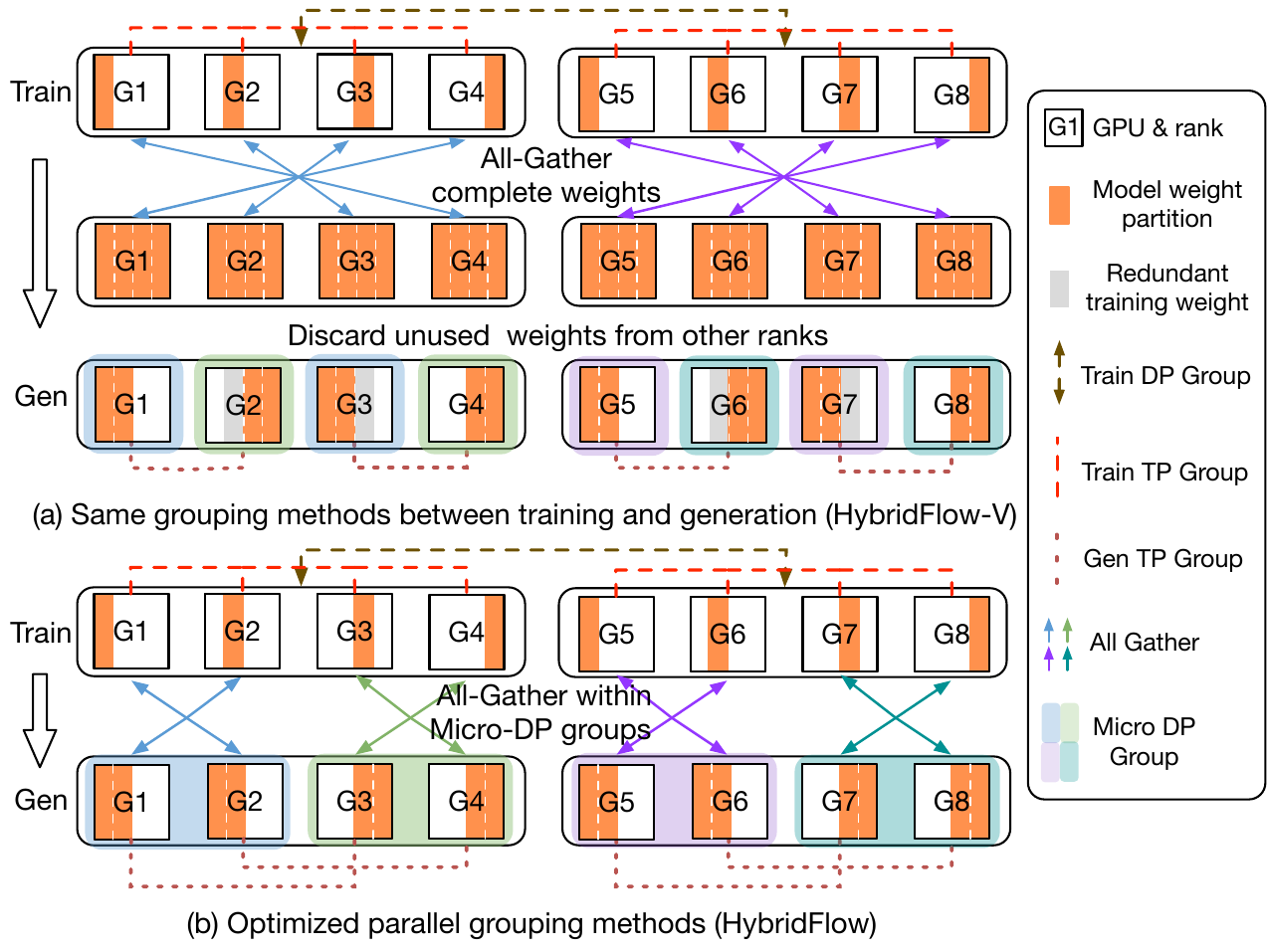}
    \vspace{-5mm}
    \caption{Model weights resharding. 2 machines each with 4 GPUs are used for actor training and generation. %
    }
    \label{fig:comm_naive}
    \vspace{-3mm}
\end{figure}

\vspace{-1mm}
\subsection{Zero redundancy model resharding} \label{sec:hybrid_zero_redundancy}
Parallel grouping methods in 3D parallelism are typically as follows: %
PP and TP groups are formed by assigning consecutive ranks to pipeline stages and tensor shards, respectively; DP groups are constructed by selecting ranks at regular intervals, determined by the product of PP size and TP size. 
In Figure~\ref{fig:comm_naive}(a), actor training uses 3D parallel groups, 1-4-2: there is one PP group for all GPUs (for illustration clarify);
the TP groups are [G1, G2, G3, G4], [G5, G6, G7, G8], and the DP groups are [G1, G5], [G2, G6], [G3, G7], [G4, G8]. %
Suppose the same parallel grouping methods are used but with different parallel sizes, e.g., 1-2-2-2 for generation in Figure~\ref{fig:comm_naive}(a).
During the transition from training to generation, 3D-HybridEngine applies all-gather operations among the model 
parallel groups to aggregate all parameters, and then retain only a subset of model weights on each device for its generation, according to the parallel groups the device belongs to.
On some GPUs (e.g., G2, G3, G6, G7), there is no overlap between training and generation model weights, and separate memory is needed to maintain weights for subsequent training as well (grey boxes in Figure~\ref{fig:comm_naive}(a)).%
We call the system \sysname-V, when 3D-HybridEngine uses the above {vanilla} parallel grouping methods in 
{the two stages.}

We further design a new parallel grouping method for 3D-HybridEngine to use in the generation stage, that eliminates the redundancy in weights storage and leads to minimal memory footprint and communication due to actor model resharding between training and generation. %
Specifically, we form generation TP and PP groups by selecting ranks at regular intervals, determined by $\frac{t}{t_g}$ and $\frac{p}{p_g}$, and construct micro DP groups by sequentially assigning ranks along the generation TP or PP dimensions.
In Figure~\ref{fig:comm_naive}(b), 1-2-2-2 parallel groups are used in generation: the generation TP groups are [G1, G3], [G2, G4], [G5, G7], [G6, G8]; %
and the micro DP groups are [G1, G2], [G3, G4], [G5, G6], [G7, G8].
This strategic rearrangement of generation parallel groups leads to overlap between training and generation model weights on each device, enabling reuse of training weights during generation and
\textit{zero redundancy} in device memory usage due to model resharding. In addition, 3D-HybridEngine conducts several all-gather operations concurrently, one within each micro DP group, leading to significantly reduced 
communication overhead.

\begin{table}
\caption{%
Transition overhead between training \& generation
}
\vspace{-2mm}
\resizebox{\linewidth}{!}{
\begin{tabular}{c|c|c|c}
\toprule
& DS-Chat& \sysname-V & \sysname \\ 
\midrule
\makecell{Comm. Vol} & $\frac{tpd - 1}{tpd}M$ & $\frac{tp - 1}{tp}M$ & $\frac{tp-t_gp_g}{t_gp_gtp}M$ \\ 
\midrule
\makecell{Peak Mem.}   & $M$   & $M$  & $\frac{1}{t_gp_g}M$ \\ \midrule
Redundancy& $\frac{1}{tpd}M$ & $\frac{1}{tp}M$ & 0 \\ \bottomrule                
\end{tabular}
}
\label{tab:comm_mem_analysis}
\vspace{-3mm}
\end{table}
\subsection{Transition overhead} \label{sec:hybrid_comm_mem}

In Table~\ref{tab:comm_mem_analysis}, we compare communication overhead and memory footprint during the transition between training and generation stages, among different actor engine designs. %
We assume model size of the actor is $M$ and $N_a$ GPUs are used for its training and generation. The actor engine in DeepSpeed-Chat conducts an all-gather operation across all GPUs during transition; \sysname-V performs this all-gather within training TP and PP groups. The communication volumes for these operations are %
$\frac{N_a-1}{N_a}M=\frac{tpd - 1}{tpd}M$ for DeepSpeed-Chat %
 and $\frac{tp-1}{tp}M$ for \sysname-V, calculated following~\cite{chan2007collective}. 
Both engines aggregate all model parameters in each GPU's memory before subsequently partitioning model states according to the generation parallel groups, resulting in a peak memory usage of model parameters $M$. %
As they cannot reuse training weights during generation on some GPUs, training weights need to be maintained on them, %
amounting to $\frac{1}{tpd}$ and $\frac{1}{tp}$ redundant memory consumption, respectively.

With our parallel grouping method for the generation stage, \sysname{} confines the all-gather operation within each micro DP group. 
The communication overhead is reduced to $\frac{d_g - 1}{tp}M = \frac{tp-t_gp_g}{t_gp_gtp}M$. %
Each GPU only needs to collect remote parameters within its micro DP group and can reuse the training weights in generation. Therefore, the peak memory usage of model parameters in \sysname{} precisely matches the model partition size on each GPU in generation, eliminating any redundancy in %
GPU memory usage.

\vspace{-3mm}
\section{Auto Device Mapping} \label{sec:auto_mapping}

Our hybrid programming model %
requires users to input the following configurations, which are referred to as a \textit{mapping} of the RLHF dataflow to the given devices: %
(a) device placement of the models in the dataflow; (b) the corresponding parallelism strategy for running each model in each stage. %

We provide an efficient %
algorithm (Algorithm~\ref{alg:mapping})  
for users to identify the optimized mapping of executing the RLHF dataflow on a given cluster of devices, that minimizes the end-to-end latency of each RLHF iteration. Given a dataflow $D$, we first explore all possible placement plans $\mathcal{P}$ %
for the models in the given cluster (Line~\ref{line:get_placement}).
For example, the PPO algorithm involves four models, resulting in 15 possible placements (from the Bell partition problem~\cite{bell1934exponential, rota1964number}),
ranging from a completely standalone placement where all models are placed on different devices (e.g., OpenRLHF's placement) to colocating all models on the same set of devices (e.g., DeepSpeed-Chat's placement). We refer to %
colocated models on the same set of GPUs as a colocated \textit{set}. Models in a colocated set can employ different parallelism strategies across the same set of GPUs. We identify the smallest number of GPUs to be allocated to each of the colocated model sets, $A_{min}$, based on memory consumption of colocated models, ensuring no out-of-memory errors (Line~\ref{line:get_min_alloc}).

Next, starting from the minimal GPU allocation in $A_{min}$, we enumerate all feasible device allocations to each colocated model set (Lines~\ref{line:enum_alloc}-\ref{line:enum_all_set}).
Given device allocation $A$ to the colocated set and computation workload $W$ of models in the set, we explore optimized parallelism strategies for each model in the \verb|auto_parallel| module, that minimizes model execution latency. 
The workload $W$ includes input and output shapes and computation (training, inference or generation) of each model. 
In \verb|auto_parallel|,  we utilize a simulator module \verb|simu| to estimate the latency of different parallel strategies, following previous research~\cite{zhongDistServeDisaggregatingPrefill2024, zheng2022alpa, yuan2024llmrooftline, llm-analysis} (outline in 
Appendix.~\ref{appendix:auto_parallel}).

The \verb|d_cost| module estimates the end-to-end latency of the RLHF dataflow under given model placement and %
parallelism strategies, by iterating through all stages in the dataflow graph and summing up latencies of all stages (Lines~\ref{line: d_cost_call}, \ref{line:d_cost}). %
For models in the same colocated set and involving computation in the same stage (such as actor and critic both performing model update in RLHF training stage),
their execution latencies are summed up (Line~\ref{alg:d_cost_sum_up}). For models in different colocated sets, their execution within the same stage can be parallelized, and the latency of the stage is determined by the maximum execution time among different sets (Line~\ref{alg:d_cost_max}). 
We identify the best device placement of the models with their corresponding parallelism strategies, achieving minimal execution time per RLHF iteration 
(Lines~\ref{line:start_compare_best}-\ref{line:end_compare_best}).

\begin{algorithm}[t]
\small
\caption{Device Mapping for an RLHF Dataflow}
\label{alg:mapping}
\begin{algorithmic}[1]
\STATE {\bfseries Input:} RLHF dataflow graph $D$, LLMs in RLHF dataflow $L$=$[l_1, l_2, \ldots, l_k]$,
workload $W$ of LLMs in RLHF dataflow, total \# of GPUs $N$,
memory capacity per GPU $Q$ 
\STATE {\bfseries Output:} device mapping of models in RLHF dataflow
\STATE $\mathcal{P} \leftarrow \text{get\_placements}(D, L, N)$ \label{line:get_placement}
\STATE $C^{*} \leftarrow \infty$
\STATE $best\_mapping \leftarrow \emptyset$
\FORALL{$plm \in \mathcal{P}$}
    \STATE $C_{plm} \leftarrow \infty$
    \STATE $best\_plm\_alloc \leftarrow \emptyset$
    \STATE $A_{min} \leftarrow \text{get\_min\_alloc}(plm, Q, N)$ \label{line:get_min_alloc}
    \FORALL{$A \in \text{enum\_alloc}(N, A_{min})$} \label{line:enum_alloc}
        \STATE $\widehat{L} \leftarrow []$ 
        \FORALL{$\text{set} \in plm$} \label{line:enum_all_set}
            \FORALL{$l \in \text{set}$}
                \STATE $\widehat{l} \leftarrow \text{auto\_parallel}(A, A_{min}, l, W)$ \label{line:auto_parallel}
                \STATE $\widehat{L}.\text{append}(\widehat{l})$
            \ENDFOR
        \ENDFOR
        \STATE $plm.\text{update}(\widehat{L})$ 
        \STATE $C_{alloc} \leftarrow \text{d\_cost}(D, plm, W)$ \label{line: d_cost_call}
        \IF{$C_{alloc} < C_{plm}$} \label{line:start_compare_best}
            \STATE $C_{plm} \leftarrow C_{alloc}$
            \STATE $best\_plm\_alloc \leftarrow (plm, A)$
        \ENDIF
    \ENDFOR
    \IF{$C_{plm} < C^{*}$}
        \STATE $C^{*} \leftarrow C_{plm}$
        \STATE $best\_mapping \leftarrow best\_plm\_alloc$ \label{line:end_compare_best}
    \ENDIF
\ENDFOR
\RETURN $best\_mapping$

\vspace{0.5mm}
\STATE {\bfseries Procedure} d\_cost($D$, $plm$, $W$): \label{line:d_cost}
\STATE $\quad s \gets \text{number of stages in } D$
\STATE $\quad c \gets [0] \times s$ 
\textit{ // Initialize latency for each stage to 0} 
\STATE $\quad \textbf{for all}\  \text{set} \in plm \ \textbf{do}$
    \STATE $\quad \quad c_{g} \gets [0] \times s$
    \STATE $\quad \quad \textbf{for all}\  i \in \{0, ..., s-1\} \ \textbf{do}$
        \STATE $\quad \quad \quad \textbf{for all}\  \widehat{l} \in \text{set} \ \textbf{do}$
            \STATE $\quad \quad \quad \quad c_g[i] \gets  c_g[i] + \text{simu}(\widehat{l}, W[i])$
            \label{alg:d_cost_sum_up}
    \STATE $\quad \quad \quad c[i] \gets  max\{c[i], c_g[i]\}$ \label{alg:d_cost_max}
\STATE $\quad \textbf{return} \  \text{sum}(c)$ 

\end{algorithmic}
\end{algorithm}

The complexity of Algorithm~\ref{alg:mapping} 
is $O(\frac{(N-1)!}{(k-1)! (N-k)!})$, where $k$ is the number of models in the dataflow and $N$ is the total number of devices to run the dataflow.
This is the worst-case complexity for enumerating all possible device allocations for a placement strategy (i.e., the standalone placement), calculated by assigning $N$ devices to $k$ models (known as the integer partition problem~\cite{andrews2004integer}). 
For better efficiency, we cache parallelism strategies identified for each model on a number of devices $A$, to eliminate redundant searches for the same parallelism strategies when the model is placed on different sets of $A$ GPUs in different placement strategies. %

Though we assume $N$ homogeneous GPUs when running the auto mapping algorithm, Algorithm \ref{alg:mapping} can be readily extended for optimizing model mapping over heterogeneous devices, by considering heterogeneous devices in \verb|simu| and \verb|auto_parallel| modules~\cite{zhang2024hap}.

\section{Implementation} \label{sec:impl}
\sysname{} is implemented in around 12k lines of Python code (LoC). %

\noindent \textbf{Hybrid programming model.}
The hierarchical APIs are implemented with 1.8k LoC. %
The centralized single controller is built on top of Ray~\cite{moritz2018ray} and uses Remote Process Calls (RPC) to coordinate the execution order of different models and transfer data between models following the dataflow.
These intermediate data are stored in TensorDict~\cite{paszke2019pytorch}.
{In our multi-controller paradigm for distributed computation, each model function runs on a separate process across various devices, with control messages relayed from each controller's CPU process to the corresponding GPU.}
Our implementation supports Megatron-LM, PyTorch FSDP, and DeepSpeed as the LLM training and inference engines, and vLLM for auto-regressive generation. %
In vLLM, we replace the centralized KVCache manager with a distributed manager to align with the multi-controller paradigm.

\noindent \textbf{3D-HybridEngine.} Its main logic %
is implemented with 2.4k LoC on top of Megatron-LM and vLLM. We %
store actor model weights for training and generation stages on separate memory buffers, offload generation weights to the CPU memory during training, reload generation weights back to GPU memory during the transition, and use both buffers in generation. 
We %
use NCCL communication primitives~\cite{jeaugey2017nccl} to collect and concatenate model parameters in each micro DP group during the transition between training and generation. We offload KVCache to CPU memory after generation and reload it back to GPU in the next iteration. %

\noindent\textbf{Auto-Mapping Algorithm}
is implemented with 1.9k LoC, together with three simulators for training, inference, and generation workloads. The algorithm is run before starting the RLHF dataflow on CPU, to generate device mapping and parallelism strategies for dataflow initialization.

\begin{figure*}[t]
\subfigure[7B (1.68$\times$$\sim$8.63$\times$)]{
\includegraphics[width=0.25\linewidth]{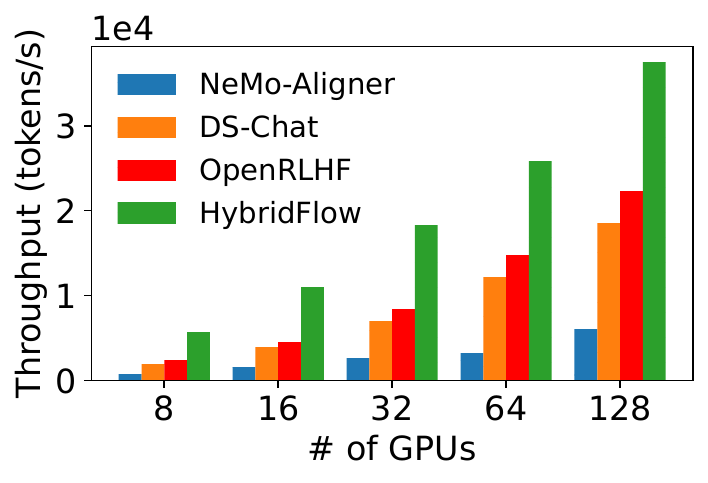}
}
\subfigure[13B (2.70$\times$$\sim$18.96$\times$)]{
\includegraphics[width=0.24\linewidth]{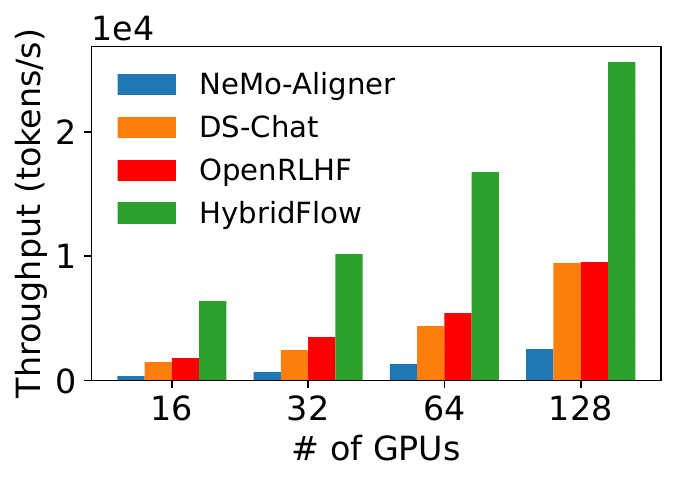}
}
\hspace{-2mm}
\subfigure[34B (2.41$\times$$\sim$20.57$\times$)]{
\includegraphics[width=0.23\linewidth]{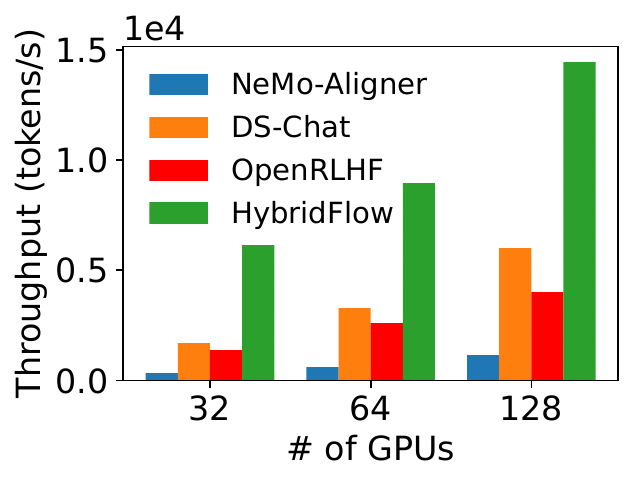}
}
\hspace{-2mm}
\subfigure[70B (5.17$\times$$\sim$17.98$\times$)]{
\includegraphics[width=0.22\linewidth]{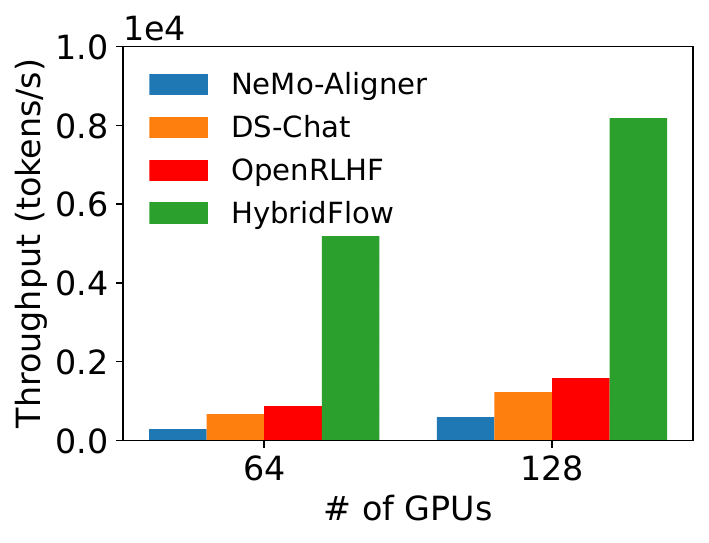}
}
\vspace{-4mm}
\caption{PPO throughput. Numbers in parentheses are \sysname{} speedups compared with baselines.}
\vspace{-5mm}
\label{fig:train_throughput}
\end{figure*}

\begin{figure*}[t]
\vspace{-1mm}
\subfigure[7B (1.53$\times$$\sim$2.56$\times$)]{
\includegraphics[width=0.25\linewidth]{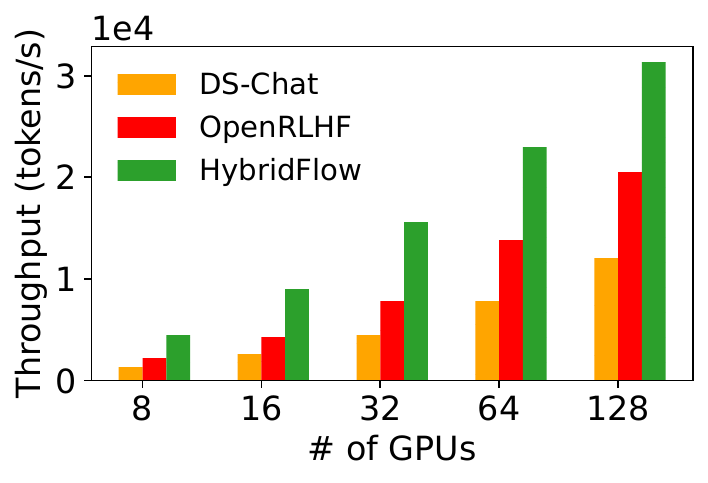}
}
\hspace{-2mm}
\subfigure[13B (2.49$\times$$\sim$3.66$\times$)]{
\includegraphics[width=0.25\linewidth]{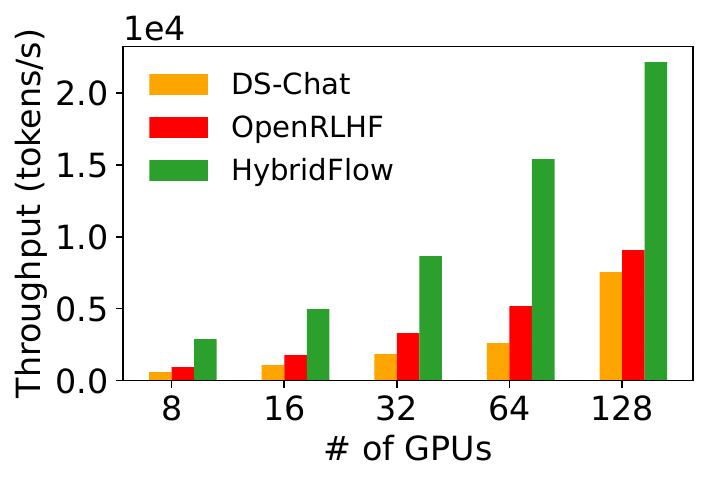}
}
\hspace{-2mm}
\subfigure[34B (2.14$\times$$\sim$4.80$\times$)]{
\includegraphics[width=0.25\linewidth]{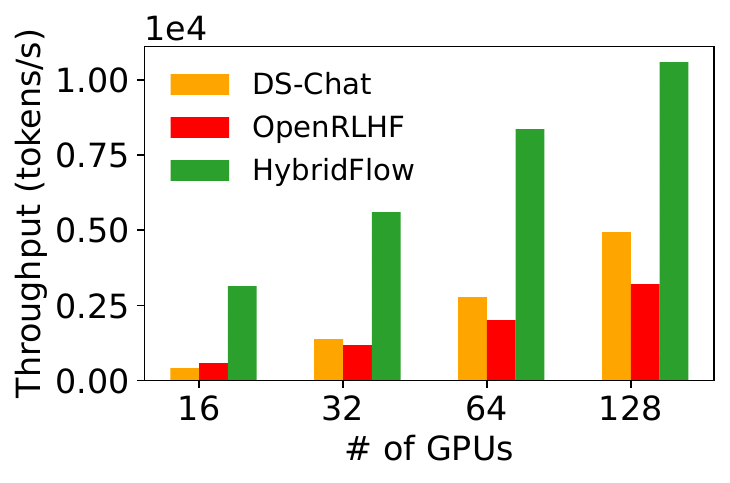}
}
\hspace{-2mm}
\subfigure[70B (6.46$\times$$\sim$9.78$\times$)]{
\includegraphics[width=0.21\linewidth]{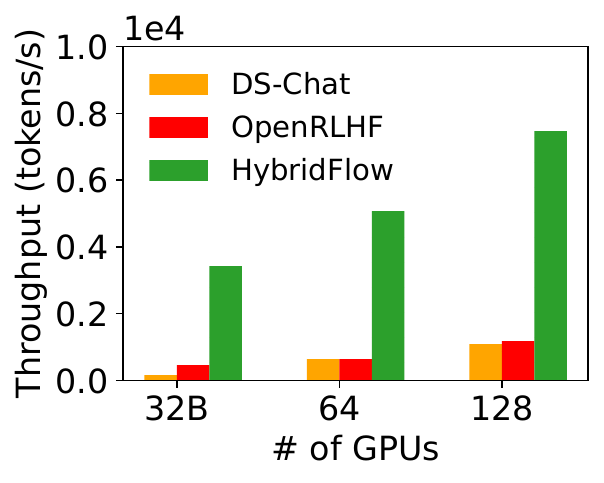}
}
\vspace{-5mm}
\caption{ReMax throughput. Numbers in parentheses are \sysname{} speedups compared with baselines}
\vspace{-5mm}
\label{fig:remax_train_throughput}
\end{figure*}

\begin{figure*}[t]
\subfigure[7B (1.71$\times$$\sim$12.87$\times$)]{
\includegraphics[width=0.25\linewidth]{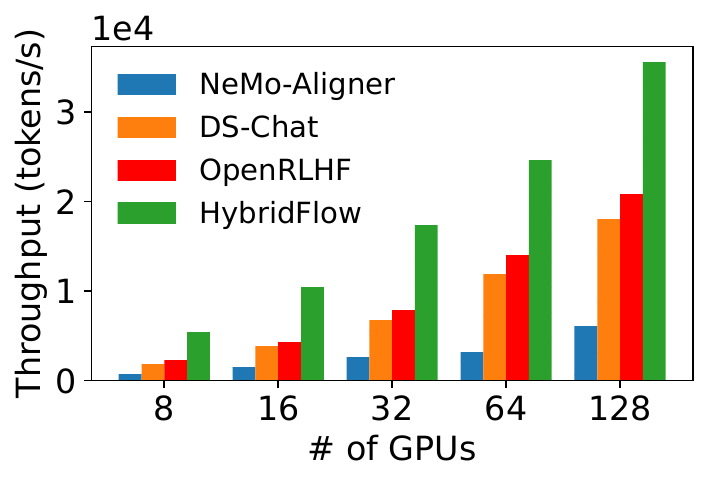}
}
\subfigure[13B (2.49$\times$$\sim$18.47$\times$)]{
\includegraphics[width=0.25\linewidth]{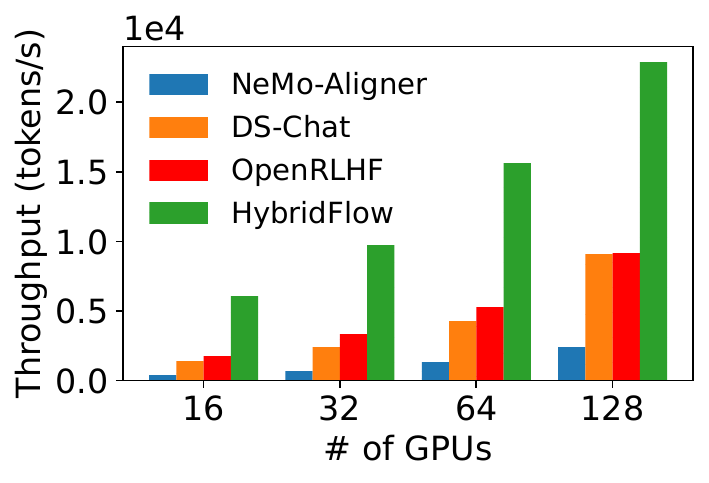}
}
\hspace{-2mm}
\subfigure[34B (2.20$\times$$\sim$19.76$\times$)]{
\includegraphics[width=0.22\linewidth]{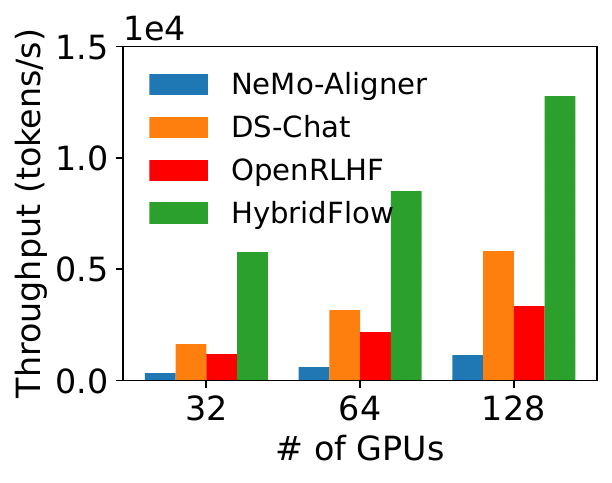}
}
\hspace{-2mm}
\subfigure[70B (4.89$\times$$\sim$16.86$\times$)]{
\includegraphics[width=0.22\linewidth]{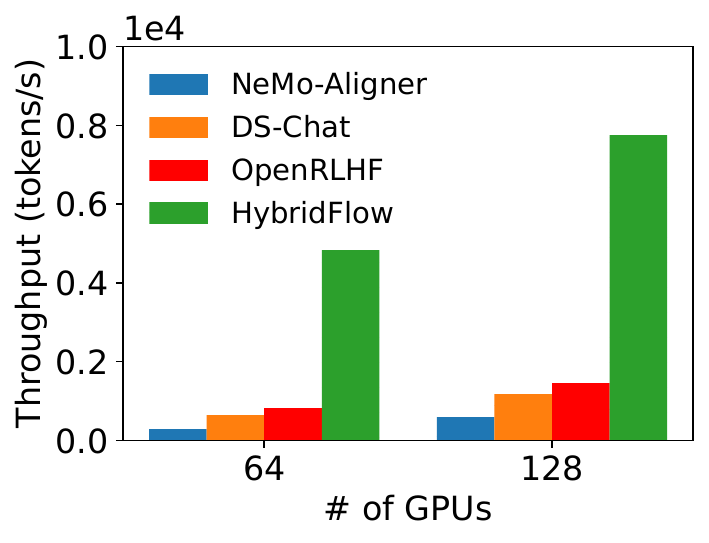}
}
\vspace{-5mm}
\caption{Safe-RLHF throughput. Numbers in the parentheses are \sysname{} speedups compared with the baselines}
\vspace{-3mm}
\label{fig:safe_rlhf_train_throughput}
\end{figure*}

\section{Evaluation}

\subsection{Experimental Setup}
\textbf{Testbed.} We deploy \sysname{} on a cluster of 16 machines (128 GPUs). Each machine is equipped with 8 NVIDIA A100-80GB GPUs inter-connected with 600GB/s NVLink. The inter-machine bandwidth is 200Gbps. Our experiments use the following software versions: CUDA12.1, PyTorch 2.1.2, Megatron-core 0.6.0, NCCL 2.18.1, and vLLM 0.3.1.

\noindent\textbf{Models and RLHF algorithms.}
We %
run the RLHF dataflow (Figure~\ref{fig:rlhf_dataflow}) of PPO~\cite{schulman2017proximal}, ReMax~\cite{li2023remax} and Safe-RLHF~\cite{daiSafeRLHFSafe2023} algorithms.
PPO is one of the most popular %
algorithms for RLHF~\cite{bai2022training, ouyang2022training}, consisting of %
actor, critic, reference policy, and reward models. Each model is a Llama~\cite{touvron2023llama} model with sizes ranging from 7B to 70B. 
Safe-RLHF has an additional cost model whose architecture and size are the same as the reward model and ReMax eliminates the critic model. 
We use mixed precision for actor and critic training, i.e., BF16 for model parameters and FP32 for gradient and optimizer states, with Adam~\cite{kingma2017adam} optimizer in all experiments. 
BF16 is used in model inference and auto-regressive generation.
If not specified, the experiment results are obtained from PPO.

\noindent\textbf{Baselines.}
We compare \sysname{} with state-of-the-art RLHF systems including DeepSpeed-Chat~\cite{yao2023deepspeedchat} v0.14.0, OpenRLHF~\cite{hu23openrlhf} v0.2.5, and NeMo-Aligner~\cite{NeMoAligner} v0.2.0 (detailed in Table \ref{tab:table_with_images}). 
NeMo-Alginer doesn't support ReMax algorithm.
We do not compare \sysname{} to other frameworks such as Trlx~\cite{havrilla2023trlx}, HuggingFaceDDP~\cite{wolf2019huggingfaces}, and Collosal-Chat~\cite{CollosalChat} as they are less representative and slower than the above baselines (as reported in ~\cite{yao2023deepspeedchat}). 

We use RLHF %
throughput (tokens/sec) as the performance metric, computed by dividing the total number of tokens in prompts and responses in a global batch by one RLHF iteration time.
All reported performance %
numbers are averaged over 5 training iterations after a warm-up of 10 iterations.

\noindent\textbf{Datasets and hyperparameters.}
We perform RLHF on "Dahoas/ful-hh-rlhf" dataset~\cite{bai2022training} of HuggingFace, which is widely used for LLM alignment%
~\cite{yuan2023rrhf, santacroce2023efficient-rlhf}. %
As the baseline systems may not incorporate continuous-batching optimization~\cite{yu2022orca} during generation, for a fair comparison, we enforce the same length on all responses to be generated. In each experiment, the input prompt length and the output response length are both 1024 and the global batch size of input prompts to the actor model is 1024.
The number of PPO epochs is 1 and the number of PPO update iterations per epoch is 8, aligning with previous RLHF research~\cite{ouyang2022training, huang2024n+, xu2024dpo}.

\begin{figure*}[t]
\begin{minipage}{0.65\textwidth}
\centering
\subfigure[13B]{
    \includegraphics[width=5.5cm]{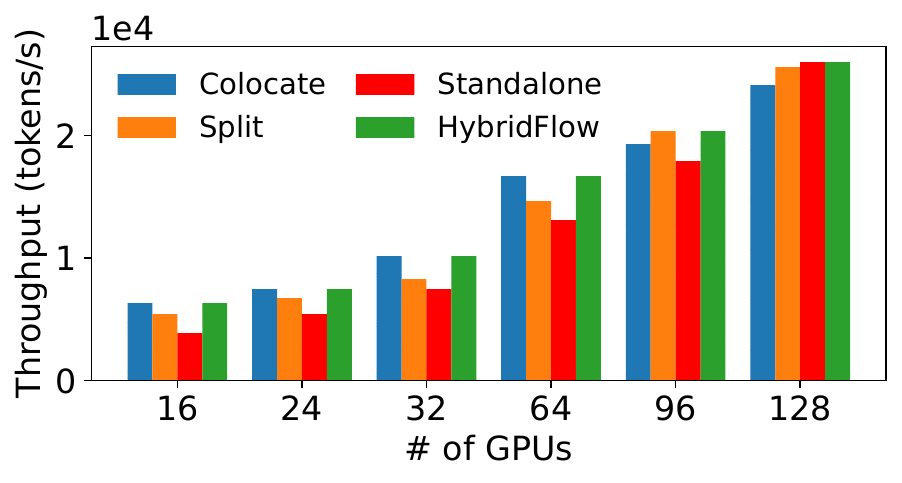}
}
\subfigure[34B]{
    \includegraphics[width=5.5cm]{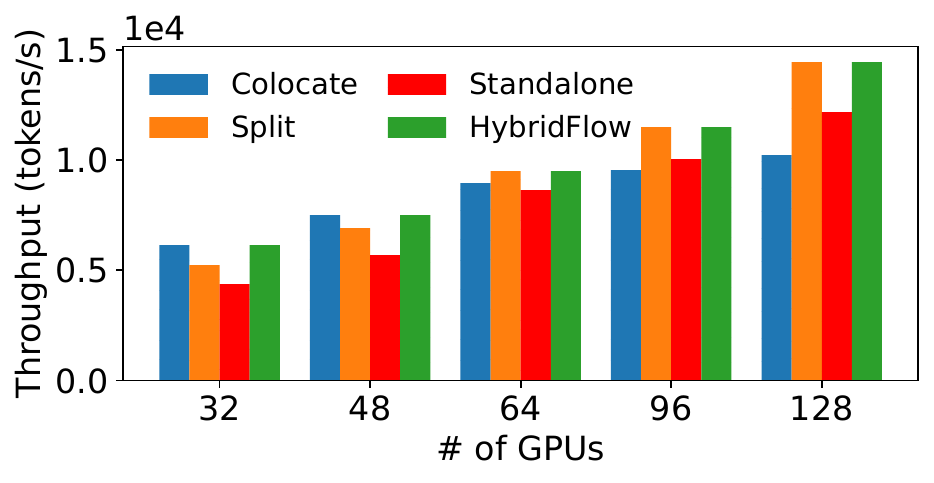}
}
\vspace{-4mm}
\caption{Throughput of \sysname{} under different placements}
\label{fig:exp_placement}
\end{minipage}
\vspace{-3mm}
\begin{minipage}{0.33\textwidth}
    \vspace{3mm}
    \includegraphics[width=5.5cm]{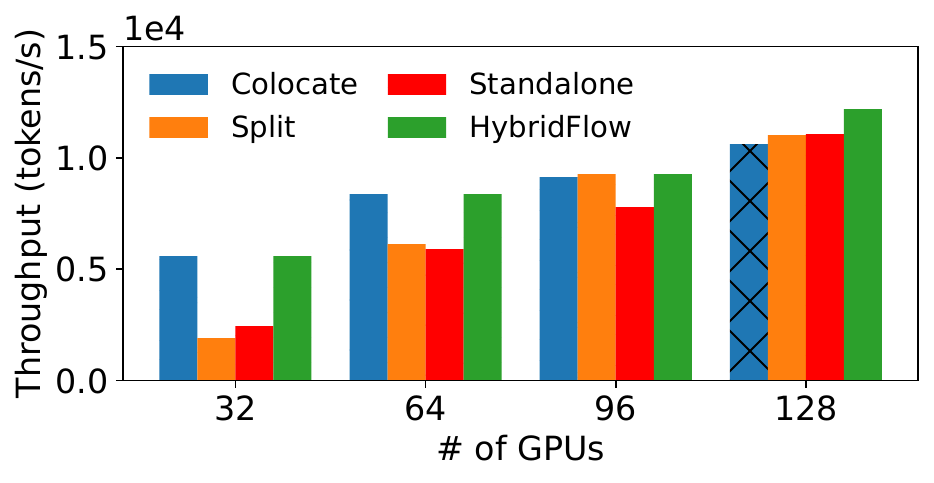}
    \vspace{-5mm}
    \caption{Placement comparison under 13B actor and reference policy \& 70B critic and reward model.}
    \label{fig:exp_mix_placement}
\end{minipage}
\end{figure*}

\subsection{End-to-End %
performance} \label{sec:exp_e2e_train_performance}
Figures~\ref{fig:train_throughput}, \ref{fig:remax_train_throughput}, and \ref{fig:safe_rlhf_train_throughput} show RLHF throughput %
when running PPO, ReMax, and Safe-RLHF respectively. 
The actor, critic, reference, and reward models in this set of experiments are of the same size, following previous practice~\cite{bai2022training, yao2023deepspeedchat, ouyang2022training}. The number of GPUs used in experiments of different model sizes ranges from the smallest number of GPUs to run RLHF without OOM
to 128 GPUs. 
We do not enable offloading optimizer states~\cite{ren2021zerooffload} in the experiments for fair comparison.

\noindent \textbf{Overall performance.}
We observe that \sysname{} consistently outperforms the baselines across all model scales. In Figure~\ref{fig:train_throughput} for PPO, \sysname{} outperforms DeepSpeed-Chat, OpenRLHF and NeMo-Aligner by 3.67$\times$ (up to 7.84$\times$), 3.25$\times$ (up to 5.93$\times$) and 12.52$\times$ (up to 20.57$\times$), respectively. This is mainly because \sysname{} effectively executes generation, inference, and training in all RLHF stages by sharding the models with different parallelism strategies to fit various computation workloads. \sysname{} achieves the highest average speedup of 9.64$\times$ when training 70B models, 
as \sysname{} reduces the transition overhead by up to 71.2\% and 89.1\% compared to DeepSpeed-Chat and OpenRLHF, which also incurs large inter-machine communication when training with ZeRO-3.
Due to the lack of KVCache in generation engine, NeMo-Aligner's main performance bottleneck lies in the generation stage, which accounts for up to 81.2\% of its %
RLHF iteration time. %
Similar %
results can be observed in Figures ~\ref{fig:remax_train_throughput}, 
\ref{fig:safe_rlhf_train_throughput}
validating the efficiency of \sysname{} on running various RLHF algorithms.

\noindent \textbf{Scalability.} %
\sysname{} achieves at least 2.09$\times$ speedup %
on 8 GPUs. With increasing GPUs, the strong scaling efficiency of \sysname{} on various model scales is 66.8\%, computed by dividing $\frac{\mbox{throughput in largest scale}}{\mbox{throughput in smallest scale}}$ by $\frac{\mbox{max. \# of GPUs}}{\mbox{min. \# of GPUs}}$%
~\cite{amdahl1967strongscaling}, averaging over three algorithms and all model scales.
Scaling to a large number of GPUs with a fixed global batch size results in smaller local batch sizes for each worker, potentially causing GPU underutilization. Running 7B models on 128 GPUs, \sysname{} still outperforms the best baseline OpenRLHF for 1.68$\times$, 1.53$\times$, and 1.71$\times$ on PPO, ReMax, and Safe-RLHF respectively. This can be attributed to HybridFlow's ability to adapt the best placement strategies for different models and cluster sizes to minimize RLHF time. OpenRLHF performs better in a larger GPU cluster but less efficiently on smaller ones.

\begin{figure*}[t]
\vspace{-2mm}
\subfigure[7B ($T_g$=2, $P_g$=1, $T$=8,$P$=1)]{
\includegraphics[width=0.25\linewidth]{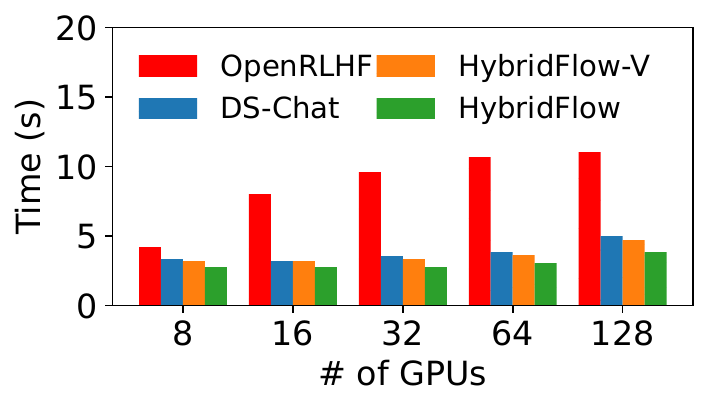}
}
\subfigure[13B ($T_g$=4, $P_g$=1, $T$=8,$P$=1)]{
\includegraphics[width=0.25\linewidth]{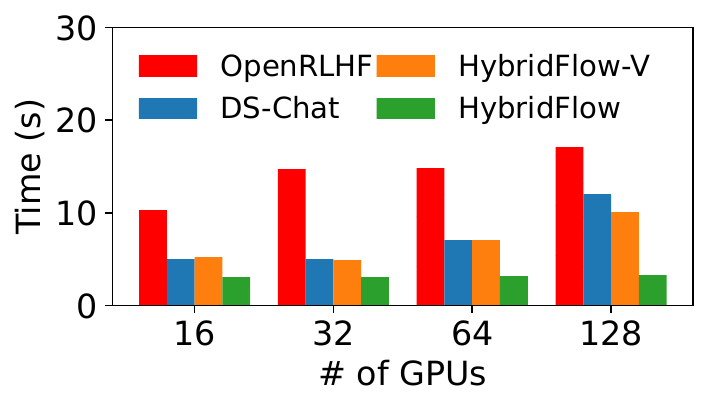}
}
\hspace{-2mm}
\subfigure[34B ($T_g$=8, $P_g$=1, $T$=8,$P$=4)]{
\includegraphics[width=0.23\linewidth]{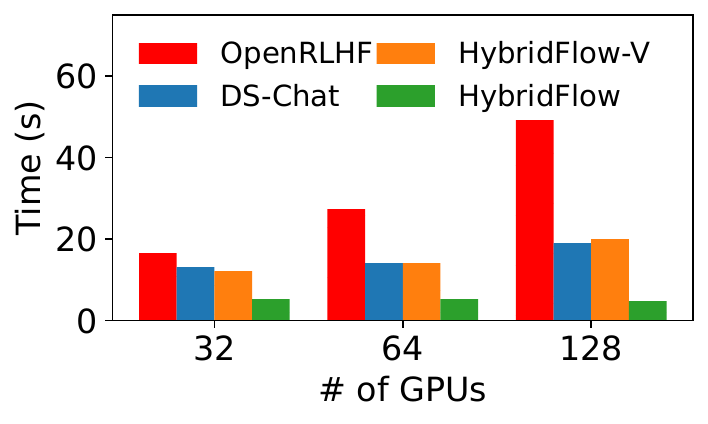}
}
\hspace{-2mm}
\subfigure[70B ($T_g$=8, $P_g$=1, $T$=8,$P$=8)]{
\includegraphics[width=0.225\linewidth]{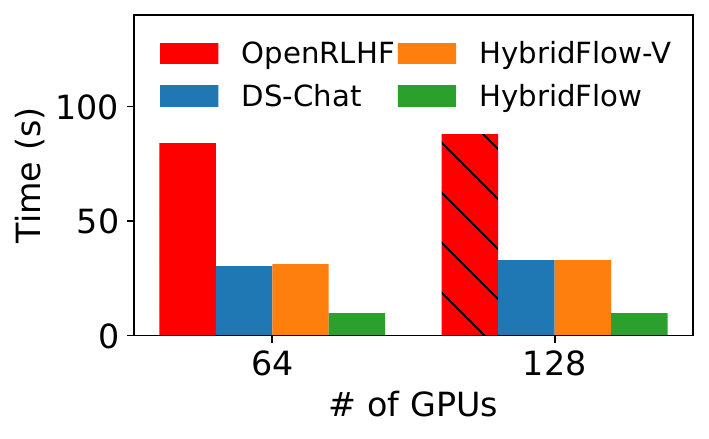}
}
\vspace{-3mm}
\caption{Transition time between actor training and generation. %
}
\vspace{-3mm}
\label{fig:exp_transit_time}
\end{figure*}

\begin{figure}[t]
    \centering
\subfigure[7B]{
    \includegraphics[width=0.48\linewidth]{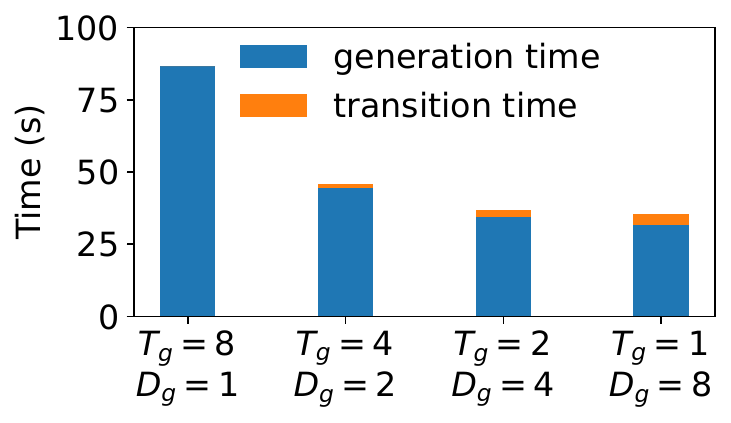}
}
\hspace{-3mm}
\subfigure[13B]{
    \includegraphics[width=0.48\linewidth]{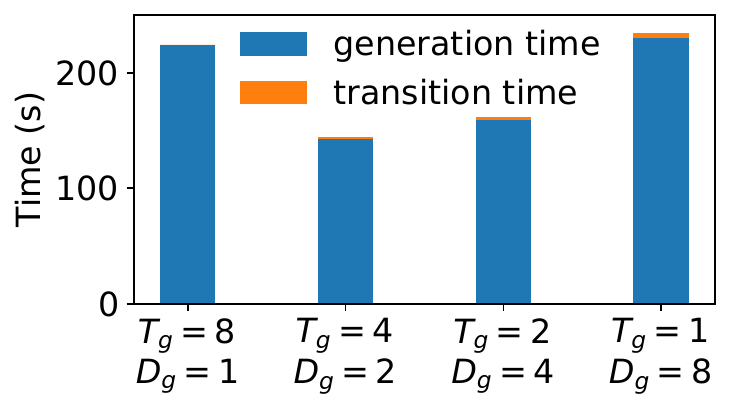}
}
    \vspace{-3mm}
    \caption{%
    Time breakdown on different generation parallel sizes of the actor model on 16 GPUs. %
    }
    \vspace{-4.5mm}
    \label{fig:hybrid_breakdown}
\end{figure}

\vspace{-2mm}
\subsection{Model Placement}%
\label{sec:exp_placement}
In this experiment, we implement various model placements of the PPO algorithm in \sysname{}, under the same model and cluster settings as in Sec.~\ref{sec:exp_e2e_train_performance}: %
(i) \textit{colocate}, the placement strategy in DeepSpeed-Chat; (ii) \textit{standalone}, that in OpenRLHF and; (iii) \textit{split}, NeMo-Aligner's colocation placement (actor and reference policy on the same set of devices and critic and reward model on another); %
(iv) \textit{hybridflow}, %
the optimized placement obtained by Algorithm \ref{alg:mapping}. %

\noindent \textbf{Comparison of different model placements.}  Figure~\ref{fig:exp_placement} reveals that optimized placement of \sysname{} under different numbers of GPUs varies. From 16 to 64 GPUs, colocating all models on the same set of devices yields the best performance. 
For 96 to 128 GPUs with 34B models and 96 GPUs with 13B models, the split strategy becomes optimal. 
The split strategy divides GPUs evenly between the two sets of models, as their sizes are equal.
For 13B models on 128 GPUs, the standalone strategy achieves the highest throughput. %
In this case, \sysname{} allocates 64 GPUs for the actor, 32 for the critic, and 16 each for the reference and reward model.
In smaller clusters, computation of all models can fully utilize GPU resources; the colocate strategy ensures maximum GPU usage in different RLHF stages.
In larger clusters, RLHF throughput under colocate placement fails to scale up linearly as the batch size is fixed and the computation-to-communication ratio decreases with a larger DP size on more GPUs.
Standalone and split strategies place models on different devices with a smaller DP size for each model in larger clusters, facilitating parallel execution of different models in the same stages. %
In all cases, our Algorithm \ref{alg:mapping} produces the best placement with the highest training throughput.

\noindent \textbf{Larger critic and reward model.} We further evaluate model placements when running PPO with a 13B actor and reference policy and 70B critic and reward models (larger critic and reward models are expected to produce better alignment~\cite{bai2022training}). Figure~\ref{fig:exp_mix_placement} shows that the colocate strategy still outperforms others by 44.8\% on average %
with up to 64 GPUs. The split strategy achieves higher throughput with 96 GPUs. When scaling to 128 GPUs, the best placement obtained by Algorithm \ref{alg:mapping} colocates actor, reference, and reward models on 64 GPUs while allocating the remaining 64 GPUs to critic.
On the same number of GPUs, actor and reference policy's computation time is much smaller than critic and reward model, and colocating the reward model with actor and reference policy reduces the GPU idle time in the experience preparation stage. 
In general, distributing actor and critic on different devices for parallel execution in the training stage leads to higher throughput in large clusters.

\subsection{%
3D-HybridEngine} \label{sec:exp_benefit_hybrid_engine}
\noindent \textbf{Transition time comparison.} 
Figure~\ref{fig:exp_transit_time} shows the transition time between actor training and generation stages on various model scales, which is the time to reshard model weights from training to generation, 
under the same settings in \textsection\ref{sec:exp_e2e_train_performance}.
OpenRLHF's transition time includes weight synchronization time between two copies of the actor model on different devices.
HybridFlow reduces the transition time by 55.2\% (11.7s) on average %
and the transition overhead by up to 89.1\% (78.2s) with 70B models, while maintaining consistent overhead across different cluster scales.
This is attributed to our new parallel grouping method for the generation stage (\textsection\ref{sec:hybrid_comm_mem}). %
In baseline methods, all model parameters must be collected during transition, necessitating layer-by-layer collections multiple times to prevent OOM.
\sysname{} enables zero memory redundancy during transition and requires only one all-gather operation per micro DP group.

\vspace{0.5mm}
\noindent \textbf{%
Transition and generation time}
We further validate the need to use different parallel sizes in actor training and generation in \sysname{}. In this experiment, all models are colocated on the same set of GPUs, and the KVCache for generation is allocated using the remaining GPU memory (i.e., best-effort allocation).
Figure~\ref{fig:hybrid_breakdown} gives the transition and generation time when running RLHF on 16 GPUs with 7B and 13B models, respectively, with training parallel groups 1-8-2 (following p-t-d convention) and varying generation TP group size $t_g$ from 1 to 8. 
The generation PP group size remains constant at $p_g$=1 and the micro DP group size $d_g$ is computed as $\frac{8}{t_g}$.
We observe that applying a smaller generation TP group size, $t_g$=2, for 7B models and $t_g$=4 for 13B models reduces the generation latency by 60.3\% and 36.4\%, respectively.
Conversely, using the same TP size as training ($t_g$=8), following the NeMo-Aligner approach, results in the largest generation latency due to GPU underutilization.
Further reducing $t_g$ fails to achieve higher speedup, as a smaller $t_g$ 
necessitates maintaining a larger KVCache per GPU.

\vspace{-1mm}
\subsection{Algorithm Runtime}
\begin{figure}[t]
    \centering
    \includegraphics[width=.95\linewidth]{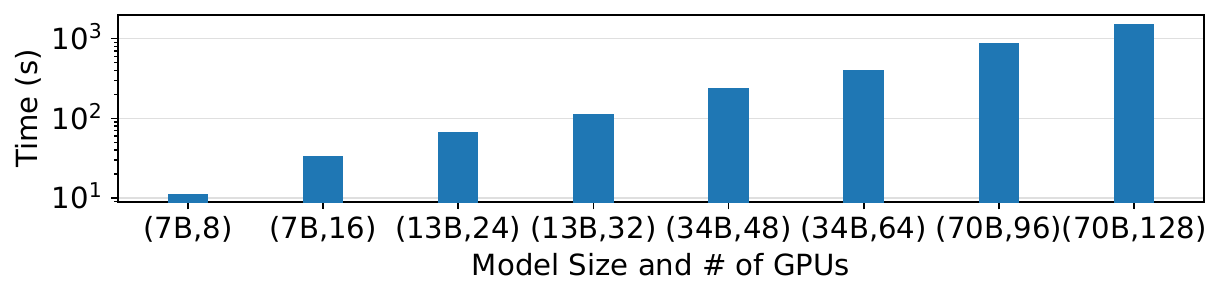}
    \vspace{-3mm}
    \caption{Runtime of device mapping algorithm. The model size and \# of GPUs are simultaneously scaled.}
    \label{fig:exp_algo_time}
    \vspace{-4mm}
\end{figure}

Figure~\ref{fig:exp_algo_time} shows the running time of Algorithm~\ref{alg:mapping}, 
which is significantly shorter than days of actual RLHF training. %
A linear growth of running time is exhibited, revealing good scalability of the device mapping algorithm with %
model size and cluster size. Most of the running time is spent on estimating the execution latency of each model's parallel strategies. More parallelism strategies are available for a larger model, requiring more simulations to identify the optimal one for each placement plan. %
Our caching of optimal parallelism strategies of the models to be reapplied across different placements reduces the search time for the best placement %
to at most half an hour.

\vspace{-1mm}
\section{Discussions} %
\noindent \textbf{Fault Tolerance.}
HybridFlow is orthogonal to existing fault-tolerance approaches~\cite{swift, wang2023gemini, jang2023oobleck, mohan2021checkfreq, eisenman2022check} and already incorporates checkpointing. Failures can be detected by NCCL errors and silent-data-corruption by checksums. Our programming model enables the single controller to coordinate checkpoint operations via RPC, allowing the saving of model states within each ParallWorker Group. 
This includes saving parameters of actor/critic models, dataloader IDs, and Random Number Generator (RNG) states to ensure system-wide consistency.
Moreover, \sysname{} can also employ redundancy-based fault-tolerance methods, such as broadcast parameters and CPU checkpoint, for fast recovery if enough healthy model replicas are available~\cite{swift, wang2023gemini}.

\noindent \textbf{Placement Insights.} We conclude three main insights for model placement and GPU allocation in RLHF training. 
\textbf{1)} Allocating more GPUs to the actor model can reduce the time-consuming generation latency, which cannot be parallelized with other models. 
\textbf{2)} When each model computation can fully utilize GPU resources, colocating all the models is most effective when training on relatively small-scale clusters.
\textbf{3)} When scaling up to large-scale clusters (i.e., strong scaling), distributing the actor and critic models on different devices for parallel execution in the training and preparation stages would help achieve higher throughput.

\noindent \textbf{Resource multiplexing.}
\sysname{} enables colocation of models on shared devices by utilizing time-sharing for GPU computation.
Recent research in DNN task scheduling has developed fine-grained resource multiplexing techniques, primarily aimed at achieving the service-level objectives of individual tasks~\cite{han2022microsecond, park2017multiplex-gpu, wang2016multiplex-gpu, liang2014multi-plexgpu, bai2020pipeswitch, han2022microsecond, cui2022dvabatch}.
Although the \texttt{ResourcePool} implementation supports parallel execution of collocated models, \sysname{} generally adheres to sequential execution to prevent GPU resource contention or OOM issues as discussed in Section~\ref{sec:2_3_rlhf_characterisitc}.
Applying GPU sharing and heterogeneous resources in RLHF training poses distinct challenges, as it seeks to balance the computation workload and manage complex data dependencies among various tasks.
Investigating fine-grained auto-mapping algorithms for GPU sharing in RLHF training, coupled with model offload optimization and integration of heterogeneous devices, would be a promising direction for future research.

\noindent \textbf{From alignment to reasoning.}
In RLHF for LLM alignment, the reward signal is generated by the reward model. Besides alignment tasks, similar algorithms (e.g., PPO and GRPO~\cite{shao2024deepseekmath}) can be applied to other domains, such as code generation and mathematical reasoning. 
For these tasks, a ground truth may exist for each prompt, which can be determined by assessing the correctness of the output value for each code test case and verifying the accuracy of mathematical results.
Therefore, the reward model can be replaced by non-neural-network reward modules, such as a sandbox environment~\cite{zhangframework} for evaluating generated code or a reward function~\cite{cobbe2021gsm8k, 2019math} to validate mathematical results. \sysname{} can seamlessly integrate these reward modules by wrapping them as remote functions and orchestrating their execution within the single-process script, providing a flexible and efficient framework for diverse reinforcement learning applications.

\section{Related Work}
\noindent\textbf{RL frameworks.} 
There have been plenty of frameworks for RL, ranging from general-purpose RL systems design for small-scale DNNs~\cite{liang2018rllib, liang2021rllib, openaibaseline, coach, hafner2017tensorflowagents, PytorchRL} to RLHF systems specifically optimized for LLMs~\cite{yao2023deepspeedchat, hu23openrlhf, NeMoAligner, xiao2023adaptive, CollosalChat}. 
We have thoroughly examined closely related work in \textsection\ref{sec:2_bckground_and_motivation} and we discuss more RL frameworks in this section. These RL frameworks~\cite{openaibaseline, coach, hafner2017tensorflowagents, PytorchRL, wang2023gear}, similar to recent RLHF systems, use a hodgepodge of multi-controller frameworks to implement their algorithms. They establish multiple long-running distributed programs with each component coordinating the execution order with hard-coded data synchronization. Gear~\cite{wang2023gear} further optimized the experience replay segment of the RL pipeline. However, all these frameworks fail to support LLM training, inference, and generation in RLHF.

\noindent\textbf{LLM training and serving systems.}
TorchDDP~\cite{paszke2019pytorch} and Horovod~\cite{sergeev2018horovod} support data parallel training. ByteScheduler~\cite{pengGenericCommunicationScheduler2019} and DeepSpeed~\cite{rasley2020deepspeed} extend data parallelism with communication and memory optimizations. 
Numerous systems~\cite{shoeybi2019megatron, jiang2024megascale, lu2017flexflow, wang2019tofu, narayanan2021efficient, fan2021dapple, zhang2022accelerating} optimized large model training through model parallelisms such as tensor parallelism and pipeline parallelism to partition models across devices. 
LLM serving systems~\cite{kwon2023efficient, agrawal2023sarathi, zhongDistServeDisaggregatingPrefill2024, yu2022orca, nvidiaTensorRTLLM, song2023powerinfer} also adopts data and model parallelism to accelerate auto-regressive generation with specialized optimizations like continuous-batching~\cite{yu2022orca} and chunked-prefill~\cite{agrawal2023sarathi}.
Note that all the above frameworks adopt multi-controller paradigm for efficient computation.

\noindent\textbf{Dataflow systems.}
Dataflow systems like MapReduce~\cite{dean2008mapreduce}, Spark~\cite{zaharia2016spark}, Dryad~\cite{isard2007dryad}, and Naiad~\cite{murray2013naiad} are popular for analytics and ML workloads but they lack support for dynamic task graphs. 
Ray~\cite{moritz2018ray} unifies task-parallel and actor programming models in a single dynamic task graph and implements a scalable distributed scheduler and a global control store, which is adopted by many RL frameworks~\cite{liang2018rllib, liang2021rllib}. 
Pathways~\cite{barham2022pathways}, a closed-source project for TPUs, are designed to easily express complex parallelism patterns and fine-grain control flow within a single DNN model, such as pipeline parallelism and Mixture-of-Experts with sparse computation. It employs an asynchronous distributed dataflow design that enables parallel control plane execution despite data dependencies, reducing the dispatch overhead from single-controller paradigm. Its main focus lies on single-model training, requiring complex compilations of each sub-network of a DNN model. \sysname{} can integrate Pathways as a submodule to implement the computation of models in the RLHF dataflow.

\vspace{-1mm}
\section{Conclusion}

HybridFlow is an RLHF framework that enables flexible representation and efficient execution of diverse RLHF algorithms. 
We propose a hybrid programming model that allows users to easily build RLHF dataflow in a few lines of code by encapsulating distributed computation of different LLMs into primitive APIs and hiding the complexity of data resharding among nodes. Our 3D-HybridEngine ensures efficient execution of training and generation of the actor model, with zero memory redundancy and significantly reduced communication overhead for model parameter resharding. Furthermore, our effective mapping algorithm optimizes GPU allocation and placement of models in the RLHF dataflow.
Extensive experiments demonstrate that HybridFlow achieves 1.53$\times$ to 20.57$\times$ speedup compared to state-of-the-art RLHF systems under various model sizes and cluster scales.

\vspace{-1mm}
\begin{acks}
We would like to thank our shepherd Y. Charlie Hu and the anonymous reviewers for their constructive feedback. 
We thank Xin Liu, Yangrui Chen, and Ningxin Zheng for their insightful feedback on this project.
This work was supported in part by a ByteDance Research Collaboration Project, and grants from Hong Kong RGC under the contracts HKU 17204423 and C7004-22G (CRF).
\end{acks}

\bibliographystyle{ACM-Reference-Format}
\bibliography{reference}

\appendix
\begin{table*}[htbp]
\caption{The transfer protocols in HybridFlow.}
\resizebox{\linewidth}{!}{%
\begin{tabular}{c|c|c|m{10cm}}
\toprule
Transfer Protocols & Distribute function & Collect function & \multicolumn{1}{c}{Use case}  \\ 
\midrule
\texttt{ONE\_TO\_ALL}  & Broadcast the data to all ranks.      & Gather the data from all ranks. & All the worker methods have the same input and run the ssme codes, e.g. model initialization.\\
\midrule
\texttt{3D\_PROTO}  & \makecell{Split the data, scatter across all DP ranks \\ and broadcast within the group.}      & \makecell{ Gather 
 and concatenate the data from\\  the \textit{p=-1, t=0} worker in all DP groups.}  & The model is sharded among multiple workers within each data-parallel group. The output of the model only exists in the last pipeline stage and is duplicated across the data-parallel groups. This is a typical scenario in 3D parallel training in Megatron-LM, Deepspeed, etc.\\
\midrule
\texttt{3D\_ALL\_MICRO\_DP}  & \makecell{Split the data by micro DP size, scatter across \\ all micro DP groups and broadcast \\ among all ranks within the group.}      & \makecell{ Gather 
 and concatenate the data from\\  the local\_rank=0 worker in all micro DP groups.}  & Used with HybridEngine. It is used to handle the 3D-parallel scheme of the policy model, when switching between training and inference.\\
\midrule
\texttt{3D\_PP\_ONLY}  & \makecell{Broadcast the data to all ranks.}      & \makecell{ Gather 
 and concatenate the data from\\  the \textit{t=0, d=0} worker in all PP groups.}  & Used to examine weight names as they are identical in TP and DP groups. \\
\midrule
\texttt{DP\_PROTO}  & \makecell{Split the data into batches and\\ scatter across all DP ranks.}      & \makecell{ Gather 
 and concatenate \\the data from all DP ranks.}  & Training model in data-parallel mode.\\
 \midrule
\texttt{ALL\_TO\_ALL}  & \makecell{No operation.}      & \makecell{ Gather the data from all ranks.}  & Used when debugging. Users can manually define the inputs of each worker and examine their outputs respectively.\\

\bottomrule
\end{tabular}}
\label{tab:transfer_proto}
\end{table*}

\section{Primitive APIs in HybridFlow} \label{appendix:primitive_apis}
In HybridFlow, we implemented the primitive of each model in RLHF training by inheriting the \texttt{3DParallelWorker}, \texttt{FSDP} \texttt{Worker} and \texttt{ZeROWorker}. The functions of these model classes are designed to decouple the distributed computation code and provide fundamental operations in RLHF for the users. This primitive design is compatible with the auto-regressive generation, forward pass, backward pass, and model update operations in the existing distributed inference and training frameworks. Users can easily customize the RLHF training dataflow (by adapting the numerical computation in the provided functions) according to the algorithm's design and benefit from reusing the underlying distributed computation implementation. We illustrate the meaning and the actual computations of these APIs in Table~\ref{tab:primitive_apis}.

\section{Transfer Protocols} \label{appendix:transfer_protocols}
We implemented transfer protocols that cover all common use cases of data resharding between models in RLHF dataflow. 
Users can utilize these pre-defined protocols to generate any RLHF dataflow. Moreover, 
Users can easily define their own transfer protocols by implementing a collect function and a distribute function. Transfer protocols decoupled the complicated data resharding and distributed training. We denote \textit{p, t, d} as the rank of the worker in pipeline-, tensor- and data-parallel group respectively. We illustrate these predefined protocols in Table~\ref{tab:transfer_proto}.
\begin{table*}[htbp]
\caption{Key functions provided in each model class. The users can use these provided functions to construct various RLHF algorithms in a few lines of code.}
\resizebox{\linewidth}{!}{%
\begin{tabular}{c|c|c|m{10cm}}
\toprule
Model & APIs & Computation & \multicolumn{1}{c}{Interpretation}  \\ 
\midrule
\multirow{3}{*}{Actor}  & \texttt{generate\_sequence}      & \makecell{auto-regressive\\generation} & Based on a batch of prompts, the actor model generates a batch of responses and returns the log probability of each token in the responses. \\
\cmidrule{2-4}
& \texttt{compute\_log\_prob}     & a forward pass   & The actor model computes the log probability of each token in the prompts and responses. This log probability is the same as the return log probability when performing generation using the same model precision. (Optional in PPO)  \\
\cmidrule{2-4}
& \texttt{compute\_loss}     & a forward pass   & The actor model computes the pretrain loss based on the pertaining dataset~\cite{bai2022training, daiSafeRLHFSafe2023, ouyang2022training}.  \\
\cmidrule{2-4}
& \texttt{update\_actor}  & \makecell{a forward, backward pass\\and model update} & Based on the advantages, returns (calculated from \texttt{compute\_advantage}) and pertaining loss, the actor model calculate the training loss and update its weights. We implement various loss for diverse RLHF algorithms including PPO~\cite{ouyang2022training}, Safe-RLHF~\cite{daiSafeRLHFSafe2023}, ReMax~\cite{li2023remax}, GRPO~\cite{shao2024deepseekmath} and others. \\
\midrule
\multirow{2}{*}{Critic} & \texttt{compute\_values}         & a forward pass & The critic model computes the values for each prompt and response.\\
\cmidrule{2-4}
& \texttt{update\_critic}  & \makecell{a forward, backward pass\\and model update} & Based on the values and returns, the critic computes a squared-error loss to update its weights. We also implement critic loss for diverse RLHF algorithms including PPO~\cite{ouyang2022training}, Safe-RLHF~\cite{daiSafeRLHFSafe2023}, ReMax~\cite{li2023remax}, GRPO~\cite{shao2024deepseekmath} and others. \\
\midrule
\makecell{Reference\\Policy}  & \texttt{compute\_ref\_log\_prob} & a forward pass  & The reference model computes the reference log probability of each token in the prompts and responses. This log probability is utilized as a benchmark to evaluate the divergence of the actor model and constrain its learning process. \\ 
\midrule
Reward  & \texttt{compute\_reward} & a forward pass      & The reward model conducts forward computation to calculate scores for a given set of prompts and responses. The rewards could be token-level or sample-level. \\
\midrule
- & \texttt{compute\_advantage}  & \makecell{numerical\\computation}    &  Based on the values rewards from the value model and reward model respectively, the function estimates the advantages on the given prompts and the current policy model's responses. This computation involves no model forward passes. \\ 

\bottomrule
\end{tabular}}
\label{tab:primitive_apis}
\end{table*}

\begin{algorithm}[htbp]
\caption{Auto Parallelism Algorithm}
\label{alg:auto_parallel}
\begin{algorithmic}[1]
\STATE {\bfseries Input:} Device allocation $A$, minimal device allocation and model parallel size for each model in a set $A_{min}$, workload $W$, the number of GPUs per machine $U$

\STATE {\bfseries Output:} the parallelism strategy for the model in a set

\STATE {\bfseries Procedure} auto\_parallel($A$, $A_{min}$, $l$, $W$):
    \STATE $N_l = A[l]$ \textit{ // Get device allocation of the model}
    \STATE $t_{min} = A_{min}[l].t$ \textit{ // Get minimal model parallel size}
    \STATE $p_{min} = A_{min}[l].p$
    \STATE $\text{best\_para} \leftarrow \emptyset$
    \STATE $\text{best\_para.cost} \leftarrow \infty$
    \FORALL{$\text{t} \in \{ t_{min}, t_{min} + 1 ..., U$\}}
        \FORALL{$\text{p} \in \{ p_{min}, p_{min} + 1 ..., \frac{N_l}{U}$\}} 
            \STATE $\text{d} \leftarrow \frac{N_l}{\text{p} \times \text{t}}$
            \STATE $\text{para\_plan} \leftarrow (p, t, d)$ 
            \STATE $\text{cost} \leftarrow \text{simu}({para\_plan}, l, W[l])$
            \IF{$\text{best\_para}.cost > \text{cost}$}
                \STATE $\text{best\_para}.cost \leftarrow \text{cost}$
                \STATE $\text{best\_para} \leftarrow \text{para\_plan}$
            \ENDIF
        \ENDFOR
    \ENDFOR
\RETURN $\text{best\_para}$

\end{algorithmic}
\end{algorithm}

\section{Auto-Parallelism Algorithm}  \label{appendix:auto_parallel}

Algorithm~\ref{alg:auto_parallel} outlines the search process of the optimal parallelism strategy of each model.
Starting from the minimal model parallelism size of each model (to prevent OOM when colocating with multiple workers), we enumerate all feasible parallel configurations based on the number of GPUs and the number of GPUs per machine $U$. The default number of $U$ is set to 8. We use \texttt{simu} module to estimate the latency of each model based on their workload. This module includes three simulators for training, inference, and generation workload, all are analytical models following previous research~\cite{yuan2024llmrooftline, zhongDistServeDisaggregatingPrefill2024, llm-analysis}. The training and inference workload is compute-bound while the generation workload is memory-bound.
For the actor model, we first find the parallelism strategy for training and record the memory usage in the training stage. 
During actor generation, KVCache requirements are calculated using the batch size and max sequence length. If the model-parallel size for the generation stage cannot accommodate both parameters and KVCache, we increase it.
Then, we seek the optimal strategy with corresponding KVCache allocation by comparing the latency estimation.
Developing a comprehensive autoregressive generation simulator that accounts for variable KVCache sizes could further enhance the auto-mapping process in RLHF research.

\end{document}